\documentclass[10pt,twocolumn,letterpaper]{article}

\usepackage[pagenumbers]{cvpr} 

\usepackage{adjustbox}
\usepackage{amsmath}
\usepackage{amssymb}
\usepackage{amsthm}
\usepackage{bm}
\usepackage{booktabs}
\usepackage{colortbl}
\usepackage{graphicx}
\usepackage{makecell}
\usepackage{mathtools}
\usepackage{multirow}
\usepackage{siunitx}
\usepackage{xcolor}

\definecolor{cvprblue}{rgb}{0.21,0.49,0.74}
\usepackage[pagebackref,breaklinks,colorlinks,allcolors=cvprblue]{hyperref}

\title{FAST: Topology-Aware Frequency-Domain Distribution Matching for Coreset Selection}

\author{
    Jin Cui\textsuperscript{$*1$}, 
    Boran Zhao\textsuperscript{$*\dagger2$}, 
    Jiajun Xu\textsuperscript{$2$}, 
    Jiaqi Guo\textsuperscript{$3$}, 
    Shuo Guan\textsuperscript{$2$}, 
    Pengju Ren\textsuperscript{$1$} \\[2mm]
    \textsuperscript{$1$}State Key Laboratory of Human-Machine Hybrid Augmented Intelligence,\\
    Institute of Artificial Intelligence and Robotics, Xi'an Jiaotong University\\
    \textsuperscript{$2$}School of Software Engineering, State Key Laboratory of Human-Machine Hybrid Augmented Intelligence,\\
    Institute of Artificial Intelligence and Robotics, Xi'an Jiaotong University\\    
    \textsuperscript{$3$}School of Mathematical Sciences, Nankai University\\
    {\tt\small {andycui@stu.xjtu.edu.cn, boranzhao@xjtu.edu.cn,    mikanaffine@stu.xjtu.edu.cn} } \\
    {\tt\small {2312195@mail.nankai.edu.cn, guanshuo@stu.xjtu.edu.cn, pengjuren@xjtu.edu.cn}}
}

\begin{document}
\maketitle
{
    \renewcommand{\thefootnote}{\fnsymbol{footnote}}
    \footnotetext[1]{Equal contribution.}    
    \footnotetext[2]{Corresponding author.}   
}
\begin{abstract}
Coreset selection compresses large datasets into compact, representative subsets, reducing the energy and computational burden of training deep neural networks. Existing methods are either: (i) DNN-based, which are inherently coupled with network-specific parameters, inevitably introducing architectural bias and compromising generalization; or (ii) DNN-free, which utilize heuristics that lack rigorous theoretical guarantees for stability and accuracy. Neither approach explicitly constrains distributional equivalence of the representative subsets, largely because continuous distribution matching is broadly considered inapplicable to discrete dataset sampling. Furthermore, prevalent distribution metrics (e.g., MSE, KL, CE, and MMD) are often incapable of accurately capturing higher-order moments differences. These deficiencies lead to suboptimal coreset performance, preventing the selected coreset from being truly equivalent to the original dataset.

We propose FAST (Frequency-domain Aligned Sampling via Topology), the first DNN-free distribution-matching coreset selection framework that formulates coreset selection task as a graph-constrained optimization problem grounded in spectral graph theory and employs the Characteristic Function Distance (CFD) to capture full distributional information (i.e., all moments and intrinsic correlations) in the frequency domain. We further discover that naive CFD suffers from a “vanishing phase gradient” issue in medium and high-frequency regions; to address this, we introduce an Attenuated Phase-Decoupled CFD. Furthermore, for better convergence, we design a Progressive Discrepancy-Aware Sampling strategy that progressively schedules frequency selection from low to high. This preserves global structure before refining local details, enabling accurate matching with few frequencies while preventing overfitting. Extensive experiments demonstrate that FAST significantly outperforms state-of-the-art coreset selection methods across all evaluated benchmarks, achieving an average accuracy gain of 9.12\%. Compared to other baseline coreset methods, it reduces power consumption by 96.57\% and achieves a 2.2$\times$ average speedup even on CPU with 1.7GB of memory, underscoring its high performance and energy efficiency.
\vspace{-6mm}
\end{abstract}    
\section{Introduction}
\label{sec:introduction}

\begin{figure}
    \centering
    \includegraphics[width=0.9\linewidth]{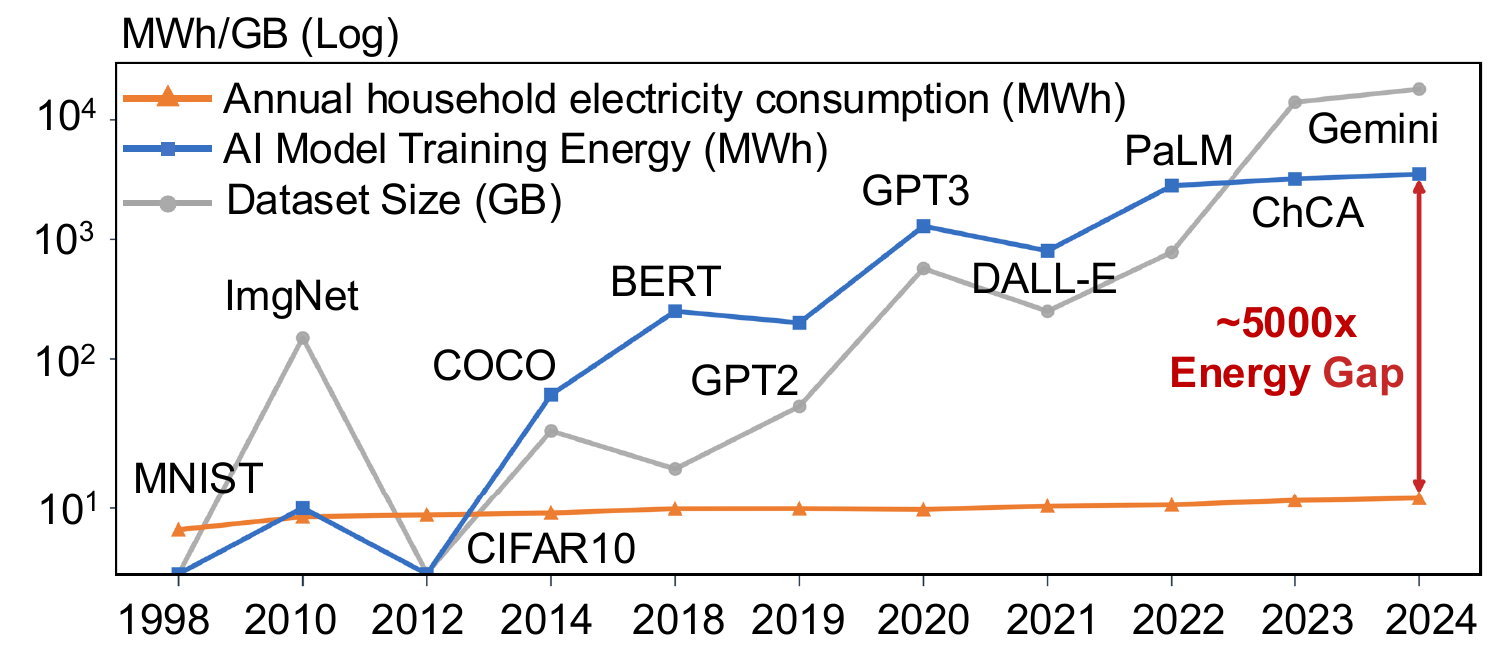}
    \vspace{-3mm}
    \caption{The energy consumption of training exceeds the annual electricity usage of numerous households by several orders of magnitude.}
    \vspace{-6mm}
    \label{fig:energy_table}
\end{figure}

\noindent Deep Neural Networks (DNNs) have achieved—and in some cases even surpassed—human-level performance across diverse domains such as vision \cite{dosovitskiy21vit, liu21swin, carion20detr}, programming \cite{chen21codex, li22alphacode, nijkamp23codegen}, and science \cite{jumper21alphafold, merchant23alab}. 
This remarkable success is primarily driven by the availability of massive training datasets \cite{russakovsky15ilsvrc, lin14coco, gao21thepile}. 
However, training on such large-scale data incurs prohibitive energy costs, as illustrated in Fig.~\ref{fig:energy_table}, the total energy consumption can reach up to $10^3$~MWh, exceeding the annual electricity usage of numerous households by several orders of magnitude. 
To mitigate this challenge, a variety of dataset compression techniques have been proposed to condense large-scale datasets into compact yet representative subsets \cite{zhao21gm, zhao23dm, wang25ncfm, iyer21submodular}. 
These methods have found widespread adoption in tasks such as neural architecture search, continual learning \cite{rebuffi17icarl, yoon22rehearsal}, and transfer learning \cite{lee24self}.

Among existing compression techniques, coreset selection \cite{killamsetty21gradmatch, mirzasoleiman20coreset, iyer21submodular} offers superior efficiency over synthesis-based distillation \cite{zhao21gm, zhao23dm, wang25ncfm} by avoiding computationally intensive nested gradient descent, making it ideal for on-device deployment. 
Furthermore, it preserves the fidelity of data and mitigates the failure modes of synthesis methods, which often struggle to generate highly discriminative samples for classification tasks with high inter-class similarity.

\begin{figure*}
    \centering
    \includegraphics[width=0.9\linewidth]{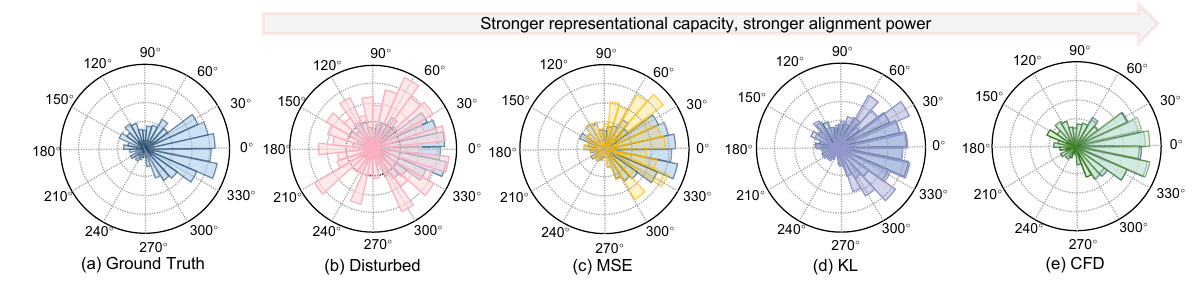}
    \vspace{-4mm}
    \caption{Comparison of distribution alignment under different metrics in frequency domain (on complex-plane). MSE aligns the mean, KL aligns both mean and variance, while CFD captures complete distributional structures in the frequency domain.}
    \vspace{-4mm}
    \label{fig:alignment compare}
\end{figure*}

 Existing coreset selection methods can be broadly categorized into two paradigms: (i) DNN-based approaches that adopt a proxy DNN to evaluate each sample’s contribution to training performance. While effective, these methods are intrinsically tied to specific network architectures, thereby introducing architectural bias and limiting generalization; and (ii) DNN-free methods that eschew DNNs entirely and rely on heuristic criteria, but often lack rigorous theoretical guarantees. 
  Critically, the efficacy of coreset selection hinges on the metrics used to identify representative samples. However, neither paradigm explicitly enforces complete distributional alignment, resulting in subsets that lack comprehensive data coverage and cannot fully represent the original dataset, thereby exhibiting instability and poor generalization. Moreover, prevalent distribution metrics such as MSE \cite{vincent08autoencoder}, KL \cite{kingma14bayes}, MMD \cite{gretton12test}, and CE \cite{krizhevsky12convnet} are constrained by the difficulty of selecting model families or kernels \cite{arjovsky17wgan} with appropriate representational capacity, rendering them incapable of capturing high-order moment differences or multivariate correlations \cite{richards17distance, cook25kld}, and ultimately leading to degraded performance, as visualized in Fig.~\ref{fig:alignment compare}.

To address these issues, recent study \cite{wang25ncfm} has introduced gradient-based distribution matching methods, which adopt continuous optimization in the feature space to improve matching accuracy. However, its application has been largely confined to synthesis-based dataset distillation, since direct continuous optimization in the feature space deviates from the discrete data manifold, preventing the discovery of corresponding real samples in the original dataset. 

To bridge this continuous-to-discrete gap, we introduce \textit{topology-aware} constraints based on spectral graph theory into the loss function, ensuring that the optimized samples maintain a one-to-one correspondence with the original data space and preserve critical local topological structures. This enables, for the first time, the application of distribution matching method with continuous optimization based on gradient descent to the discrete coreset selection task.
\begin{figure}
    \centering
    \includegraphics[width=0.9\linewidth]{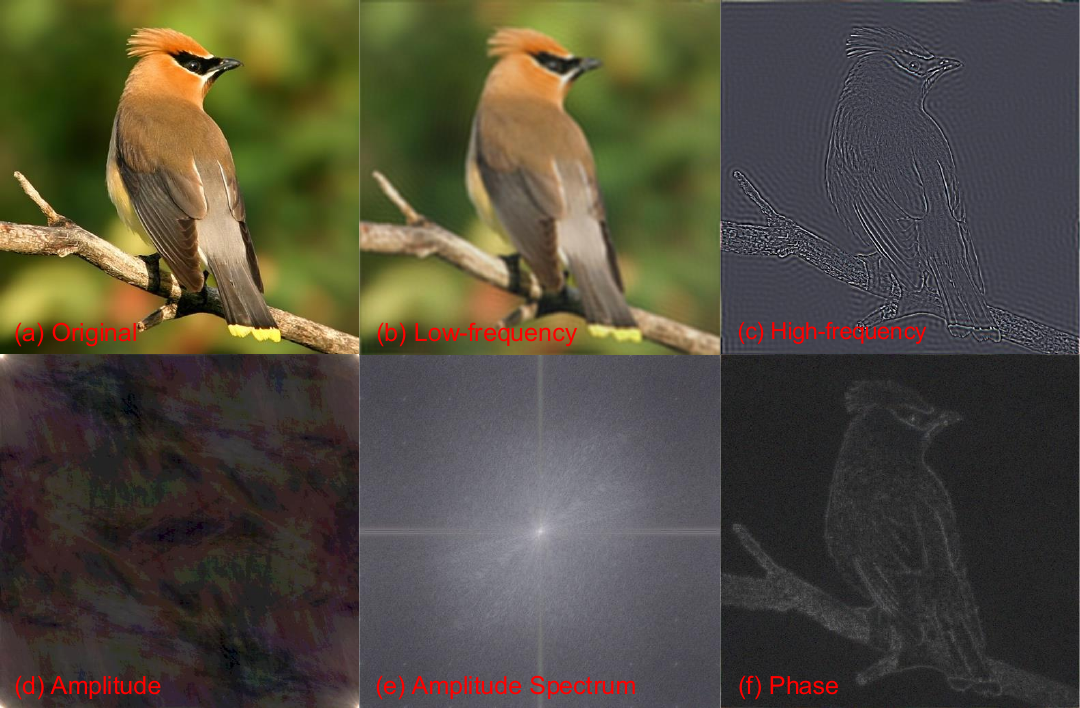}
    \caption{ Low frequencies (b) capture smooth shading and coarse shapes; high frequencies (c) capture edges and fine textures. Amplitude (d) (and corresponding spectrum (e)) encodes the energy distribution across frequencies, while phase (f) specifies the spatial arrangement of structures.}
    \vspace{-5mm}
    \label{fig:fft_decompose}
\end{figure}
Most of important, we leverage frequency-domain features as the evaluation criterion, specifically employing the Characteristic Function Distance (CFD). The underlying Characteristic Function (CF) uniquely \textit{captures all correlations and high-order moments} by projecting them into the frequency domain, thereby providing a more rigorous guarantee of distributional equivalence. Furthermore, we identify and address a critical \textit{vanishing phase gradient} problem in standard CFD, where phase–magnitude coupling blinds the optimizer to medium and high-frequency details (e.g., edges and textures, as illustrated in Fig.~\ref{fig:fft_decompose}). We propose a novel Attenuated Phase-Decoupled CFD (PD-CFD) to resolve this issue and experimentally validate its superiority on datasets rich in high-frequency content (e.g., DTD \cite{cimpoi14dtd}).

We further find that the efficacy of CFD significantly depends on frequency selection. We observe that premature inclusion of high-frequency components destabilizes optimization by causing an overemphasis on fine details while mismatching the global structure. To address this, we design Progressive Discrepancy-Aware Sampling (PDAS), a curriculum-learning strategy that matches frequencies progressively from low to high, ensuring stable and accurate alignment with a minimal set of essential frequencies.

Building upon these insights, we propose Frequency-domain Aligned Sampling via Topology (FAST) framework. FAST improves accuracy by an average of 9.12\% over state-of-the-art (SOTA) coreset selection methods, even when running on a CPU with only 1.7\,GB of memory, while reducing power consumption by 96.57\% and achieving a 2.2$\times$ speedup on average, benefited from the fast convergence realized by powerful distribuction matching of our proposed methods. Our main contributions are as follows:

\begin{enumerate}

\item We propose a novel, topology-aware, DNN-free coreset selection framework that, for the first time, enables distribution matching in the discrete domain while eliminates architectural bias.

\item We are the first to employ Characteristic Function Distance (CFD) in coreset selection to evaluate full distributional information, and we introduce Attenuated Phase-Decoupled CFD to address the \textit{vanishing phase gradient} issue in medium and high-frequency regions.

\item We conduct a thorough analysis of frequency selection and propose the curriculum-learning-based Progressive Discrepancy-Aware Sampling strategy for robust matching with a minimal number of frequencies.

\item Our method achieves SOTA performance with significantly lower computational overhead, making it suitable for deployment in resource-limited environments.

\end{enumerate}
\section{Related Work}
\label{sec:related_work}
\noindent\textbf{Dataset Condensation.}
Dataset Condensation (DC) comprises two main branches: synthesis-based distillation \cite{zhou22nfr, zhao21gm, cazenavette22mtt, zhao23dm, wang25ncfm} and sampling-based coreset selection. 
Coreset selection circumvents the prohibitive nested training overhead of distillation, offering superior efficiency, and is broadly categorized as either DNN-based or DNN-free.

DNN-based approaches typically rely on a fixed (often DNN-provided) feature embedding for geometric sampling \cite{sener18coreset, chen10herding}, or utilize training signals (e.g., losses, gradients) supplemented by adjustment mechanisms to identify informative samples \cite{paul21datadiet, moosavi16deepfool, killamsetty21gradmatch, killamsetty21glister}. 
However, this deep coupling with the proxy DNN introduces strong architectural bias, leading to poor generalization and incurring significant computational overhead. 
DNN-free methods eliminate this bias and are more efficient, but existing work is scarce, the only representative method \cite{zhao25nms} relies on manifold reduction and heuristic grid sampling—an unreliable approach lacking stable guarantees across diverse datasets.

Critically, both paradigms face inherent limitations:  
(1) Their selection metrics (e.g., MSE, KL) are insufficient to capture full distributional discrepancies.  
(2) Their guiding principles are incomplete, either emphasizing specific training contributions, compromising generalization, or relying on heuristics that match only low-order statistics, leading to instability.  
As a result, the selected coresets fail to reflect the full data distribution, ultimately degrading downstream performance.

\noindent\textbf{Characteristic Function.}
The Characteristic Function (CF) of a probability distribution (its Fourier transform) uniquely determines the distribution in the frequency domain, making the Characteristic Function Distance (CFD) an effective metric for comparing complex distributions. In statistics, it serves as a powerful non-parametric statistic for two-sample testing \cite{epps86test, fernandez08test, chwialkowski15test}. This capability has also been adopted in Generative Models, where CFD (or its variant, MMD \cite{ansari20cf}) is employed as a loss function to train GANs \cite{bruck24gnncf} by minimizing the real-to-generated discrepancy. Similarly, in domain adaptation, CFD is used as a domain discrepancy metric to learn domain-invariant representations \cite{wu20cfd}. However, to the best of our knowledge, no prior work has utilized CFD as an evaluation metric for coreset selection.

The only work applying frequency-domain features to dataset compression is NCFM \cite{wang25ncfm}, but it is limited to synthesis-based methods and thus inapplicable to discrete coreset selection. Furthermore, NCFM suffers from two critical limitations: 
(1) It relies on DNNs for feature extraction and frequency selection, thereby inheriting architectural bias, lacking interpretability, and leading to suboptimal frequency selection. 
(2) Its loss function couples phase with magnitude, failing to distinguish phase differences in high-frequency regions where magnitude decays.

\begin{figure*}[t]
    \centering
    \includegraphics[width=0.95\linewidth]{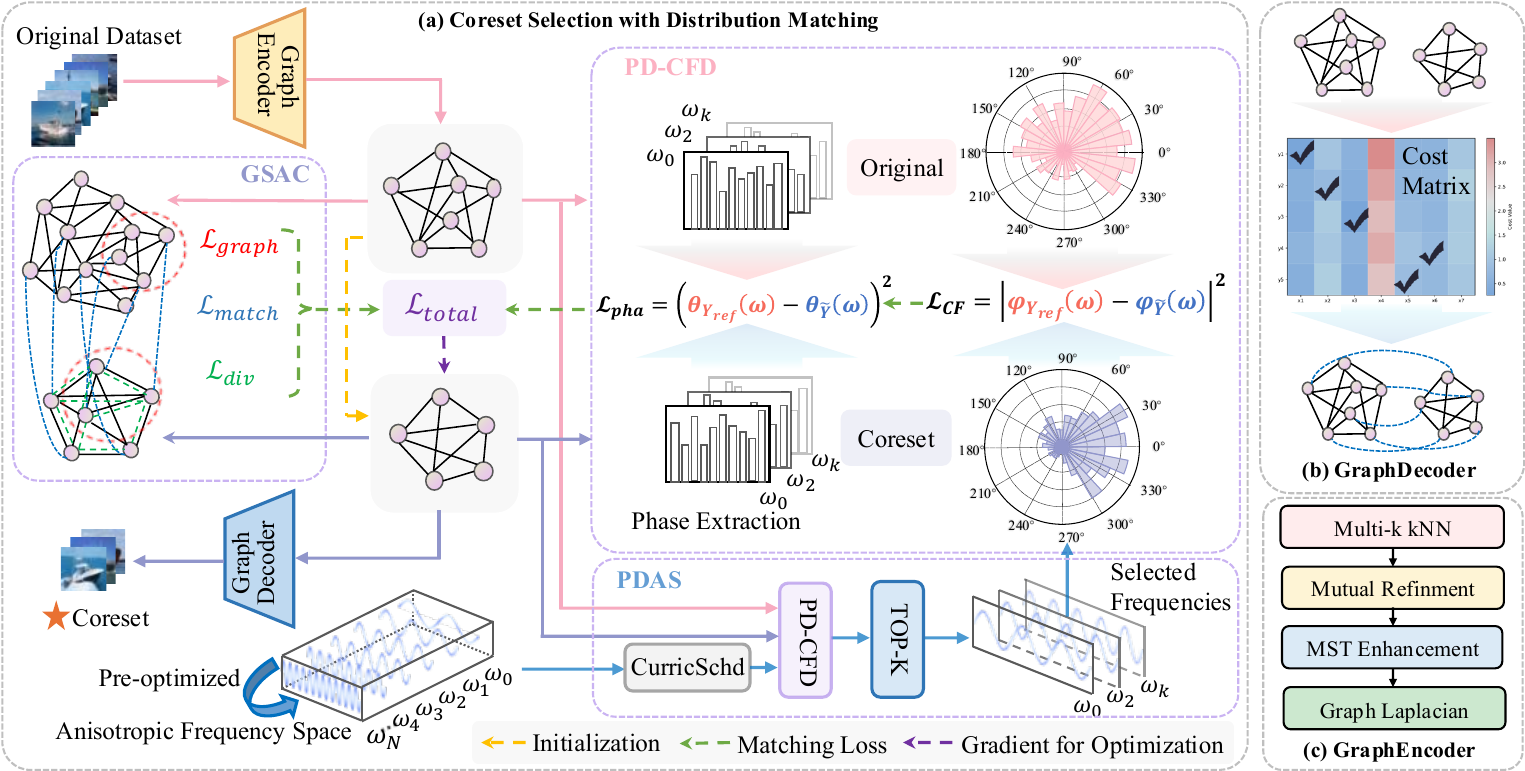}
    \vspace{-2mm}
    \caption{(a) Overview of proposed FAST. Graph-Structure-Aware Constraints (GSAC) preserves topological consistency, while Progressive Discrepancy-Aware Sampling (PDAS) progressively aligns distributions via phase-decoupled characteristic function distance (PD-CFD). (b)Graph Decoder. Maps the optimized coreset back to the original data space, ensuring structural consistency. (c) Graph Encoder. Constructs the graph topology based-on spectral graph theory.} \vspace{-4mm}
    \label{fig:placeholder}
\end{figure*}

\section{Method}
\label{sec:method}

\subsection{Graph Construction}
\noindent To provide a stable optimization space for distribution matching, we first capture the intrinsic manifold structure of the data. We construct a multi-scale weighted undirected graph $\bm{B} \in \mathbb{R}^{N \times N}$ based on the fuzzy topological theory from UMAP \cite{mcinnes18umap}, which serves as the foundation for all subsequent feature extraction and topological constraints.

\noindent\textbf{Multi-Scale Manifold Graph Construction.} To capture geometry at multiple scales, rather than relying on a single $k$-nearest neighbor scale, we construct graphs for a set of $k$-nearest neighbor scales $\{k_1, k_2, \dots\}$. For each scale $k$ and data point $\bm{x}_i$, we find the local connectivity distance $\rho_i$ and the local scale factor $\sigma_i$ by solving the following equation:
\begin{equation}
\sum_{j=1}^{k} \exp\left(\frac{-\max(0, d(\bm{x}_i, \bm{x}_{i_j}) - \rho_i)}{\sigma_i}\right) = \log_2(k)\vspace{-2mm}
\end{equation}
This defines the directed edge weights. We first symmetrize these weights for each scale $k$ using the probabilistic t-conorm (a fuzzy set union), $\bm{A}_k = \bm{A}_k + \bm{A}_k^T - \bm{A}_k \circ \bm{A}_k^T$. Subsequently, we fuse the graphs $\bm{A}_k$ from all scales into a single multi-scale adjacency matrix $\bm{B}$ by iteratively applying the same fuzzy set union. Finally, we ensure global connectivity by incorporating the graph's Minimum Spanning Tree (MST) via $\bm{B} = \bm{B} \cup \text{MST}(\bm{B})$.

\noindent\textbf{Spectral Embedding as Manifold Features.}
From the multi-scale graph $\bm{B}$, we compute its Symmetric Normalized Laplacian:
\begin{equation}
\mathcal{L}_{\text{sym}} = \bm{I} - \bm{D}^{-1/2} \bm{B} \bm{D}^{-1/2}\vspace{-2mm}
\end{equation}
where $\bm{D}$ is the diagonal degree matrix. The eigenvectors of $\mathcal{L}_{\text{sym}}$ serve as a discrete approximation of the manifold's Laplace-Beltrami operator, revealing its intrinsic geometry. We extract the $d$ eigenvectors ($d \ll N$) corresponding to the smallest non-zero eigenvalues to form the $N \times d$ manifold feature matrix $\bm{V}_{\text{full}}$. This matrix provides a robust representation of the original data's geometry and serves as the reference for all subsequent optimizations.

\subsection{Topology-Aware Alignment and Regularization}
\noindent Our method optimizes a continuous coreset representation $\tilde{Y} \in \mathbb{R}^{M \times d}$ ($M \ll N$) by aligning its frequency-domain features with the reference $\bm{V}_{\text{full}}$. The core challenge lies in the continuous-to-discrete gap: $\tilde{Y}$ must ultimately map back to $M$ discrete samples from $\bm{V}_{\text{full}}$. To address this, the optimization is guided by a composite constraint system that enforces both diversity and topological alignment.

\noindent\textbf{Diversity Constraint.}
To ensure $\tilde{Y}$ spans the feature space and covers the full distribution, we introduce a diversity loss $\mathcal{L}_{\text{div}}$ based on Determinantal Point Processes (DPP). We minimize $\mathcal{L}_{\text{div}} = -\log \det(\bm{K})$, where $\bm{K} = \bm{\Psi} \bm{\Psi}^T + \delta \bm{I}$ is the Gram matrix of the RFF features ($\bm{\Psi}$) of $\tilde{Y}$, explicitly penalizing feature redundancy.

\noindent\textbf{Graph-Aware Alignment Constraint.}
To bridge the gap between continuous optimization and discrete selection, we introduce a complementary constraint inspired by graph alignment (GUNN), which maintain structural consistency between $\tilde{Y}$ and a subset of $\bm{V}_{\text{full}}$ throughout the optimization. This constraint is composed of two main components.

At each optimization step, we find an optimal, bijective mapping $\pi: \{1,..,M\} \to \{1,..,N\}$ by solving the linear assignment problem using the Hungarian Algorithm. This mapping assigns each point $\bm{y}_i \in \tilde{Y}$ to a unique real feature $\bm{v}_j \in \bm{V}_{\text{full}}$. The assignment is guided by a graph-aware cost matrix $\bm{C} \in \mathbb{R}^{M \times N}$ whose entries consider both Euclidean proximity and topological significance:\vspace{-1mm}
\begin{equation}
C_{i,j} = \frac{||\bm{y}_i - \bm{v}_j||^2}{\text{deg}(\bm{v}_j) + \epsilon}
\vspace{-2mm}
\end{equation}
where $\text{deg}(\bm{v}_j)$ is the degree of node $\bm{v}_j$ in the original graph $\bm{B}$, anchoring points in continuous space to real nodes that are central to the manifold's topology.

Based on the computed optimal mapping $\pi$, we apply two alignment losses:
(i)  A \textit{Positional Loss ($\mathcal{L}_{\text{match}}$)} that pulls each $\bm{y}_i$ to its assigned discrete anchor $\bm{V}_{\text{full}}[\pi(i)]$:
\begin{equation}
\mathcal{L}_{\text{match}} = \frac{1}{M} \sum_{i=1}^M ||\bm{y}_i - \bm{V}_{\text{full}}[\pi(i)]||^2\vspace{-2mm}
\end{equation}
and (ii) A \textit{Topological Loss ($\mathcal{L}_{\text{graph}}$)} that preserves the original structure of the selected anchors. Let $\mathcal{L}_{\text{sub}}$ be the $M \times M$ submatrix of $\mathcal{L}_{\text{sym}}$ indexed by $\pi$. The loss is the Laplacian regularizer:
\vspace{-1mm}
\begin{equation} 
\mathcal{L}_{\text{graph}} = \text{Tr}(\tilde{Y}^T \mathcal{L}_{\text{sub}} \tilde{Y})\vspace{-1mm}
\end{equation}
$\mathcal{L}_{\text{graph}}$ ensures that if anchor points $\bm{v}_{\pi(i)}, \bm{v}_{\pi(j)}$ are strongly connected in the original manifold, their continuous counterparts $\bm{y}_i, \bm{y}_j$ also remain close in the optimization space.

\subsection{Phase-Decoupled Characteristic Function Distance}

\noindent We optimize the coreset $\tilde{Y}$ by minimizing its distance in the CF space from the reference dataset $Y_{\text{ref}}$. We build our objective function by first defining the standard CFD.

\noindent\textbf{From CF to CFD.}
The theoretical Characteristic Function (CF) of a $d$-dimensional distribution $P$ provides a complete frequency-domain representation, defined as $\varphi_P(t) = \mathbb{E}_{X \sim P}[e^{i \langle t, X \rangle}]= \int_{\mathbb{R}^d} e^{i \langle t, x \rangle} dP(x)$, where $t \in \mathbb{R}^d$ is the frequency vector. In practice, we use the Empirical Characteristic Function (ECF) computed from a set of samples $\mathcal{Y}$:
\vspace{-2mm}
\begin{equation}
\varphi_{\mathcal{Y}}(t) = \frac{1}{|\mathcal{Y}|} \sum_{y \in \mathcal{Y}} e^{i \langle t, y \rangle}\vspace{-2mm}
\end{equation}

Our goal is to match the ECF $\varphi_{\tilde{Y}}(t)$ of the coreset, to that of the reference set $\varphi_{Y_{\text{ref}}}(t)$, over a frequency distribution $t$.

The $L^2$ CFD measures the distance between two theoretical distributions $P$ and $Q$ by integrating the squared difference of their CFs over a weighting function $w(t)$:
$D^2(P, Q) = \int_{\mathbb{R}^d} |\varphi_P(t) - \varphi_Q(t)|^2 w(t) dt$. In practice, the empirical CFD (ECFD) replaces theoretical CFs with their empirical estimates and approximates the integral as an expectation over a frequency sampling distribution $p(t)$ (corresponding to $w(t)$):
\vspace{-2mm}
\begin{equation}
\mathcal{L}_{\text{CFD}} = \mathbb{E}_{t \sim p(t)}[|\varphi_{\tilde{Y}}(t) - \varphi_{Y_{\text{ref}}}(t)|^2]\vspace{-2mm}
\end{equation}

This standard ECFD serves as the foundation for our main loss $\mathcal{L}_{\text{main}}$.

\noindent\textbf{The Vanishing Phase Gradient Problem.}
To analyze the behavior of $\mathcal{D}_{w}^{2}$, we examine its integrand in polar form with magnitude $A(t) = |\varphi(t)|$ and phase $\theta_\varphi(t)$ (omitting $t$ for brevity): \vspace{-2mm}
\begin{equation} 
|\varphi_{\tilde{Y}} - \varphi_{Y_{\text{ref}}}|^2 = A_{\tilde{Y}}^2 + A_{Y_{\text{ref}}}^2 - 2 A_{\tilde{Y}} A_{Y_{\text{ref}}} \cos(\theta_{\tilde{Y}}- \theta_{Y_{\text{ref}}})   \vspace{-2mm}
\end{equation} 

This expansion exposes the \textit{Vanishing Phase Gradient} problem:  the contribution of the phase difference $\Delta \theta = (\theta_{\tilde{Y}} - \theta_{Y_{\text{ref}}})$ is scaled by (coupled with) the magnitude product $A_{\tilde{Y}} A_{Y_{\text{ref}}}$. According to the Riemann–Lebesgue lemma (see Appendix), magnitudes $A(t) \to 0$ as frequency $||t|| \to \infty$. This coupling suppresses the phase gradient in medium-high frequency regions, causing naive CFD to ignore essential information before the phase degenerates into high-frequency noise.

\noindent\textbf{Phase-Decoupled CFD Loss.}
To address this, we introduce a Phase-Decoupled loss function $\mathcal{L}_{CF}$. This loss is defined for each sampled frequency $\omega \sim p(t)$ used in the Monte Carlo estimation:  
\begin{equation} \label{eq:decoupled_loss}
\mathcal{L}_{CF}(\omega) = |\varphi_{Y_{\text{ref}}}(\omega) - \varphi_{\tilde{Y}}(\omega)|^2 + \lambda_{\phi}(\omega)(\theta_{Y_{\text{ref}}}(\omega) - \theta_{\tilde{Y}}(\omega))^2  
\end{equation}
where $\theta(\omega)$ denotes the phase angle, and $\lambda_{\phi}(\omega)$ represents a phase-penalty term. Constant penalty $\lambda_{\phi}$ is suboptimal as it amplifies high-frequency noise (where $A(t) \to 0$ and phase $\theta(t)$ is unstable). Therefore, we propose PD-CFD, using a penalty $\lambda_{\phi}(\omega)$ that adaptively decays in noisy, high-frequency regions:
\vspace{-2mm}
\begin{equation}
\quad \lambda_{\phi}(\omega) = \frac{\lambda_{p}}{1 + \alpha ||\omega||^2}\vspace{-2mm}
\end{equation}

Here, $\lambda_{p}$ and $\alpha$ control the penalty and its decay rate. This formulation amplifies the mid-range phase signal (where $A(\omega)$ has decayed but $\lambda_{\phi}(\omega)$ remains significant) while suppressing noise.
The final objective is the average of this loss over $k$ adaptively sampled frequencies $\{\omega_j\}_{j=1}^k$:
\vspace{-2mm}
\begin{equation}
\mathcal{L}_{main} = \frac{1}{k} \sum_{j=1}^k \mathcal{L}_{CF}(\omega_j)
\end{equation}

\vspace{-2mm}
\subsection{Progressive Discrepancy-Aware Sampling}
\label{sec:cf_moment_selection}
\vspace{-1mm}

\noindent\textbf{Moment Encoding in Characteristic Functions.}
The characteristic function $\varphi(t)$ provides a complete representation of a distribution by systematically encoding all of its moments. The mixed partial derivatives of $\varphi(t)$ evaluated at $t=0$ correspond directly to the mixed raw moments, $\partial^{\alpha}\varphi(0) = i^{|\alpha|}\mathbb{E}[X^{\alpha}]$. This property enables $\varphi(t)$ to be expressed through its multivariate Taylor expansion:
\vspace{-4mm}

\begin{equation}
\vspace{-1mm}
\varphi(t) = \sum_{|\alpha|\le m}\frac{1}{\alpha!}\partial^{\alpha}\varphi(0)t^{\alpha} = \sum_{|\alpha|\le m}\frac{i^{|\alpha|}}{\alpha!}\mathbb{E}[X^{\alpha}]t^{\alpha}\vspace{-1mm}
\end{equation}

This expansion reveals that the value of $\varphi(t)$ at any nonzero frequency $t$ is a weighted polynomial combination of all its constituent moments $\mathbb{E}[X^{\alpha}]$.
Furthermore, the log-CF, $\psi(t) = \log \varphi(t)$, similarly encodes the mixed cumulants $\kappa_{\alpha}$, which capture higher-order dependencies, via its derivatives at the origin: $\partial^{\alpha}\psi(0) = i^{|\alpha|}\kappa_{\alpha}$. Detailed derivations are provided in the Appendix.

The objective of the CFD is to estimate the integral measure $\mathcal{D}_{w}^{2}(P,Q)=\int |\varphi_{P}(t)-\varphi_{Q}(t)|^{2} w(t) dt$. The moment-encoding property implies that if two distributions $P$ and $Q$ differ in any $k$-th order moment, their characteristic functions $\varphi_P(t)$ and $\varphi_Q(t)$ must diverge in certain regions of the frequency domain. Therefore, the frequency sampling distribution $p(t)$ (corresponding to $w(t)$) is a critical hyperparameter that determines the estimator's sensitivity to specific types of distributional discrepancies.

\noindent\textbf{Anisotropic Frequency Initialization.}
Instead of adopting a standard Gaussian frequency bank $\mathcal{N}(0, I)$, we first perform an initialization stage to construct an anisotropic frequency space that is sensitive to the current data distribution $Y_{\text{ref}}$. This stage partitions the frequency selection range into low, medium, and high-frequency bands based on the norm $\|\omega\|$. For each band, we optimize a set of anisotropic scaling (variance) coefficients $\mathbf{s}_{\text{band}}$ by maximizing the CF difference within that band:

\vspace{-6mm}
\begin{equation}
\mathbf{s}_{\text{band}}^* = \arg\max_{\mathbf{s}} \mathbb{E}_{t \sim \mathcal{N}(0, \text{diag}(\mathbf{s}^2))} \left[ \mathcal{L}_{CF}(t) \mid t \in \text{Band} \right]
\end{equation}
This procedure assigns a data-driven anisotropic scaling to each frequency in the frequency space, enhancing sensitivity to the structural differences of the dataset. The resulting \textit{Anisotropic Frequency Library (AFL)} serves as the foundation for subsequent progressive sampling.

\noindent\textbf{Progressive Discrepancy-Aware Sampling.}
In the main optimization loop, we employ a curriculum-based strategy for adaptive frequency selection from the pre-optimized AFL. We define a frequency norm upper bound $\tau_t$ that progressively increases with iteration $t$. The candidate frequency pool at iteration $t$ is restricted to $C_t = \{\omega \in \text{AFL} \mid \|\omega\| \le \tau_t\}$. This progressive strategy ensures the model first matches low-frequency global statistics before gradually shifting to high-frequency fine-grained structures.
Within $C_t$, importance sampling is performed based on the current discrepancy. The probability $p_t(\omega)$ of sampling a frequency $\omega \in C_t$ is proportional to a composite score:\vspace{-2mm}
\begin{equation} \label{eq:sampling}
p_t(\omega) \propto \mathcal{L}_{CF}(\omega) \cdot \mathcal{D}(\omega) \hspace{5mm}(\omega \in C_t)\vspace{-1mm}
\end{equation}
where $\mathcal{L}_{CF}(\omega)$ is the total phase-decoupled loss defined in Eq.~\eqref{eq:decoupled_loss}, $\mathcal{D}(\omega)$ represents a diversity term that penalizes high correlations with previously selected frequencies. This sampling strategy ensures that each batch $\mathcal{B}_t$ for gradient estimation consistently focuses on the most informative, high-discrepancy frequencies within the current boundary $\tau_t$.

\subsection{Joint Optimization}

\noindent We guide the gradient descent optimization of the coreset $\tilde{Y}$ by minimizing a comprehensive loss function $\mathcal{L}_{total}$ that combines the primary frequency-domain matching loss with topological regularization terms:\vspace{-2mm}
\begin{equation}
    \mathcal{L}_{total} = \mathcal{L}_{main} + \lambda_{div}\mathcal{L}_{div} + \mathcal{L}_{align}\vspace{-2mm}
\end{equation}
where $\mathcal{L}_{align}$ is a composite constraint term that includes two key sub-terms from our dual alignment constraint:\vspace{-2mm}
\begin{equation}
    \mathcal{L}_{align} = \lambda_{match}\mathcal{L}_{match} + \lambda_{graph}\mathcal{L}_{graph}\vspace{-2mm}
\end{equation}
\section{Experiment}

\label{sec:experiment}
\subsection{Setup}
\noindent\textbf{Datasets.} 
We evaluate FAST on multiple benchmarks of varying scales and complexities: (1) \textit{Small-scale:} CIFAR-10 (10 classes, 60k 32$\times$32 images), CIFAR-100 (100 classes, 60k 32$\times$32 images), and SVHN (10 classes, 99k 32$\times$32 images). (2) \textit{Medium-scale:} Tiny ImageNet (200 classes, 120k 64$\times$64 images). (3) \textit{Large-scale and texture-rich:} DTD (47 classes, 5,640 images resized to 224$\times$224) and RESISC45 (45 classes, 31k 256$\times$256 images). 

\noindent\textbf{Models.}
We comprehensively evaluate the generalization of our method on a diverse range of network architectures. These include standard CNNs (ResNet18, ResNet50) as general benchmarks, lightweight CNNs (ShuffleNetV2, MobileNetV2) and a Transformer (ViT). All models are trained for 200 epochs using SGD with a momentum of 0.9, weight decay of $5\times10^{-4}$. The batch size is set to 256 for 32$\times$32 datasets and 64 for 224$\times$224 or 256$\times$256 datasets. Sampling ratios of 10\%, 20\%, and 30\% are used to assess performance under different compression levels.

\noindent\textbf{Baseline methods.}
We compare FAST against a comprehensive set of coreset selection baselines, categorized as shown in Table~\ref{tab:main result}. Among them, only FAST and NMS are DNN-free methods.

\noindent\textbf{Hardware Platform.}
Experiments are conducted on two platforms: 
(1) \textit{CPU/GPU:} Intel(R) Xeon(R) Platinum 8358P CPU, NVIDIA A100 GPU (80\,GB), and 512\,GB system memory; 
(2) \textit{Edge:} Rockchip RK3588 (4$\times$Cortex-A76 and 4$\times$Cortex-A55, NPU disabled) with 4\,GB memory. 
Energy measurements are obtained using Zeus toolkit \cite{zeus-nsdi23} and Intel’s Running Average Power Limit (RAPL) interface.

\begin{table*}[t]
    \centering
    \caption{Results of FAST on CIFAR-10/100, SVHN, TinyImageNet, and high resolution datasets DTD and RESISC-45. (Dist. M.: Distribution Matching, Grid S.: Grid Sampling, Des. B.: Desicion Boundary, Grad. M.: Gradient Matching, Bil. O.: Bilevel Optimization)}
    \vspace{-2mm}
    \begin{adjustbox}{max width=\textwidth}
    \setlength{\tabcolsep}{5pt}
    \begin{tabular}{cc|ccc|ccc|ccc|ccc|ccc|ccc}
    \toprule
    \multicolumn{1}{c|}{\multirow{2}[2]{*}{Category}} & Dataset & \multicolumn{3}{c|}{CIFAR-10} & \multicolumn{3}{c|}{CIFAR-100} & \multicolumn{3}{c|}{RESISC-45} & \multicolumn{3}{c|}{SVHN} & \multicolumn{3}{c|}{DTD} & \multicolumn{3}{c}{TinyImageNet} \\
    \multicolumn{1}{c|}{} & Ratio(\%) & 10    & 20    & 30    & 10    & 20    & 30    & 10    & 20    & 30    & 10    & 20    & 30    & 10    & 20    & 30    & 10    & 20    & 30 \\
    \midrule
    \multicolumn{1}{c|}{Dist. M.} & \cellcolor[rgb]{ .886,  .937,  .855}FAST & \cellcolor[rgb]{ .886,  .937,  .855}\textbf{90.32 } & \cellcolor[rgb]{ .886,  .937,  .855}\textbf{93.39 } & \cellcolor[rgb]{ .886,  .937,  .855}\textbf{94.93 } & \cellcolor[rgb]{ .886,  .937,  .855}\textbf{66.61 } & \cellcolor[rgb]{ .886,  .937,  .855}\textbf{72.79 } & \cellcolor[rgb]{ .886,  .937,  .855}\textbf{75.85 } & \cellcolor[rgb]{ .886,  .937,  .855}\textbf{85.00 } & \cellcolor[rgb]{ .886,  .937,  .855}\textbf{89.32 } & \cellcolor[rgb]{ .886,  .937,  .855}\textbf{91.14 } & \cellcolor[rgb]{ .886,  .937,  .855}\textbf{92.86 } & \cellcolor[rgb]{ .886,  .937,  .855}\textbf{94.53 } & \cellcolor[rgb]{ .886,  .937,  .855}\textbf{95.30 } & \cellcolor[rgb]{ .886,  .937,  .855}\textbf{45.77 } & \cellcolor[rgb]{ .886,  .937,  .855}\textbf{54.69 } & \cellcolor[rgb]{ .886,  .937,  .855}\textbf{61.85 } & \cellcolor[rgb]{ .886,  .937,  .855}\textbf{34.55 } & \cellcolor[rgb]{ .886,  .937,  .855}\textbf{51.49 } & \cellcolor[rgb]{ .886,  .937,  .855}\textbf{56.85 } \\
    \midrule
    \multicolumn{1}{c|}{Grid S.} & NMS   & 86.57  & 90.18  & 92.58  & 55.13  & 63.77  & 68.79  & 74.26  & 80.12  & 83.37  & 91.76  & 93.89  & 94.08  & 38.63  & 50.28  & 59.94  & 31.41  & 47.97  & 52.30  \\
    \midrule
    \multicolumn{1}{c|}{\multirow{2}[1]{*}{Geometry}} & Herding & 71.49  & 72.87  & 78.37  & 40.44  & 50.19  & 54.89  & 36.57  & 49.62  & 56.86  & 61.72  & 86.50  & 93.00  & 17.02  & 26.84  & 45.22  & 22.24  & 39.55  & 49.78  \\
    \multicolumn{1}{c|}{} & kCenter & 77.82  & 83.99  & 88.63  & 41.29  & 53.58  & 61.33  & 69.45  & 80.36  & 83.54  & 81.73  & 90.33  & 93.36  & 26.22  & 37.64  & 44.46  & 22.06  & 42.01  & 52.48  \\
    \midrule
    \multicolumn{1}{c|}{\multirow{3}[1]{*}{Scoring}} & Entropy & 56.37  & 68.92  & 81.68  & 25.93  & 39.49  & 49.54  & 47.18  & 65.62  & 76.38  & 69.67  & 90.19  & 94.05  & 17.25  & 29.58  & 33.82  & 14.61  & 28.12  & 41.45  \\
    \multicolumn{1}{c|}{}  & Forget & 69.07  & 80.63  & 87.18  & 48.07  & 58.32  & 65.70  & 72.64  & 79.98  & 81.42  & 77.50  & 91.81  & 93.90  & 36.73  & 50.24  & 55.33  & 30.31  & 47.84  & 53.86  \\
    \multicolumn{1}{c|}{} & GraNd & 45.46  & 62.67  & 75.48  & 19.64  & 27.89  & 39.85  & 35.72  & 55.70  & 69.25  & 50.99  & 86.47  & 92.85  & 24.60  & 37.28  & 41.33  & 10.45  & 20.07  & 30.09  \\
    \midrule
    \multicolumn{1}{c|}{\multirow{2}[2]{*}{Des. B.}} & Cal   & 81.76  & 81.88  & 85.86  & 56.04  & 63.23  & 68.12  & 64.22  & 70.36  & 73.15  & 81.19  & 87.69  & 90.19  & 40.45  & 49.72  & 58.07  & 30.92  & 47.41  & 52.86  \\
    \multicolumn{1}{c|}{} & Dfool & 59.78  & 75.19  & 81.21  & 33.47  & 43.82  & 53.20  & 57.79  & 70.16  & 76.71  & 67.33  & 87.88  & 93.07  & 24.88  & 40.10  & 48.79  & 21.74  & 40.71  & 47.85  \\
    \midrule
    \multicolumn{1}{c|}{\multirow{2}[2]{*}{Grad. M.}} & Craig & 61.21  & 70.01  & 81.24  & 36.45  & 43.17  & 53.05  & 61.23  & 74.21  & 77.02  & 65.75  & 89.63  & 93.15  & 37.18  & 46.57  & 50.90  & 24.71  & 38.05  & 45.14  \\
    \multicolumn{1}{c|}{} & GradM & 60.73  & 68.46  & 78.85  & 32.47  & 43.47  & 48.31  & 52.62  & 62.94  & 73.32  & 62.17  & 86.24  & 92.62  & 33.49  & 46.41  & 53.44  & 27.35  & 39.60  & 50.25  \\
    \midrule
    \multicolumn{1}{c|}{Bil. O.} & Glister & 57.42  & 69.45  & 79.22  & 31.24  & 40.83  & 50.52  & 53.01  & 62.84  & 72.12  & 58.70  & 86.23  & 92.36  & 34.15  & 45.81  & 49.90  & 26.07  & 39.09  & 49.50  \\
    \midrule
    \multicolumn{1}{c|}{\multirow{3}[2]{*}{Submod.}} & FL    & 78.83  & 82.55  & 86.25  & 53.62  & 58.38  & 62.99  & 76.05  & 81.27  & 83.14  & 80.92  & 90.08  & 93.22  & 40.60  & 45.46  & 44.70  & 31.75  & 45.04  & 50.99  \\
    \multicolumn{1}{c|}{} & GC    & 85.02  & 85.34  & 89.06  & 58.18  & 64.86  & 69.22  & 70.57  & 75.91  & 78.04  & 88.24  & 90.96  & 93.13  & 35.62  & 50.45  & 54.71  & 31.66  & 46.76  & 51.88  \\
    \multicolumn{1}{c|}{} & DQ    & 85.21  & 87.90  & 90.98  & 55.61  & 60.12  & 64.67  & 72.12  & 78.37  & 80.12  & 90.07  & 92.24  & 93.18  & 36.12  & 49.13  & 55.84  & 31.24  & 45.79  & 52.03  \\
    \midrule
    \multicolumn{2}{c|}{Whole Dataset} & \multicolumn{3}{c|}{95.56 } & \multicolumn{3}{c|}{80.59 } & \multicolumn{3}{c|}{94.01 } & \multicolumn{3}{c|}{96.37 } & \multicolumn{3}{c|}{71.65 } & \multicolumn{3}{c}{66.35 } \\
    \bottomrule
    \end{tabular}%
    \end{adjustbox}
    \label{tab:main result}%
    \vspace{-4mm}
\end{table*}%

\begin{table*}[htbp]
  \centering
  \caption{Comparison of computational efficiency. (FAST is CPU-only, incurring zero GPU overhead)} \vspace{-2mm}
    \begin{adjustbox}{max width=\textwidth}
    \begin{tabular}{c|cccccccccccccc}
    \toprule
    Works & \multicolumn{1}{l}{\cellcolor[rgb]{ .886,  .937,  .855}FAST-CPU} & \multicolumn{1}{l}{\cellcolor[rgb]{ .886,  .937,  .855}FAST-EDGE} & \multicolumn{1}{l}{kCenter} & \multicolumn{1}{l}{NMS} & \multicolumn{1}{l}{DQ} & \multicolumn{1}{l}{Cal} & \multicolumn{1}{l}{Craig} & \multicolumn{1}{l}{DeepFool} & \multicolumn{1}{l}{GraNd} & \multicolumn{1}{l}{Forgetting} & \multicolumn{1}{l}{GradM} & \multicolumn{1}{l}{Glister} & \multicolumn{1}{l}{Submodular} & \multicolumn{1}{l}{Uncertainty} \\
    \midrule
    Time/s & \cellcolor[rgb]{ .886,  .937,  .855}\textbf{353.0} & \cellcolor[rgb]{ .886,  .937,  .855}960.0 & 616.7 & 398.0   & 426.3 & 542.8 & 437.9 & 1079.0  & 4403.0  & 410.3 & 431.4 & 425.0   & 811.3 & 414.3 \\
    Energy/Wh & \cellcolor[rgb]{ .886,  .937,  .855}1.409 & \cellcolor[rgb]{ .886,  .937,  .855}\textbf{0.67} & 50.66 & 21.01 & 34.57 & 42.35 & 37.2  & 96.46 & 402.4 & 33.81 & 35.47 & 39.36 & 59.02 & 39.05 \\
    Accuracy & \cellcolor[rgb]{ .886,  .937,  .855}\textbf{90.32} & \cellcolor[rgb]{ .886,  .937,  .855}90.31 & 78.2  & 88.0    & 85.2  & 72.8  & 64.6  & 63.3  & 54.6  & 68.5  & 63.9  & 69.8  & 76.0    & 60.7 \\
    \bottomrule
    \end{tabular}%
    \end{adjustbox}
  \label{tab:energy}%
  \vspace{-3mm}
\end{table*}%

\subsection{Results and Analysis}
\noindent\textbf{Overall Performance.} 
We validated FAST's effectiveness across datasets of different scales and keep rates. As demonstrated in Table~\ref{tab:main result}, FAST consistently outperforms all competing baselines. It achieves an average accuracy improvement of 17.63\% over DNN-based methods and 9.12\% over SOTA DNN-free methods. Critically, FAST demonstrates a substantial 21.93\% average performance gain on DTD and RESISC45, datasets characterized by complex textures and edges (further analyzed in Appendix).

\begin{figure}
    \centering
    \includegraphics[width=0.85\linewidth]{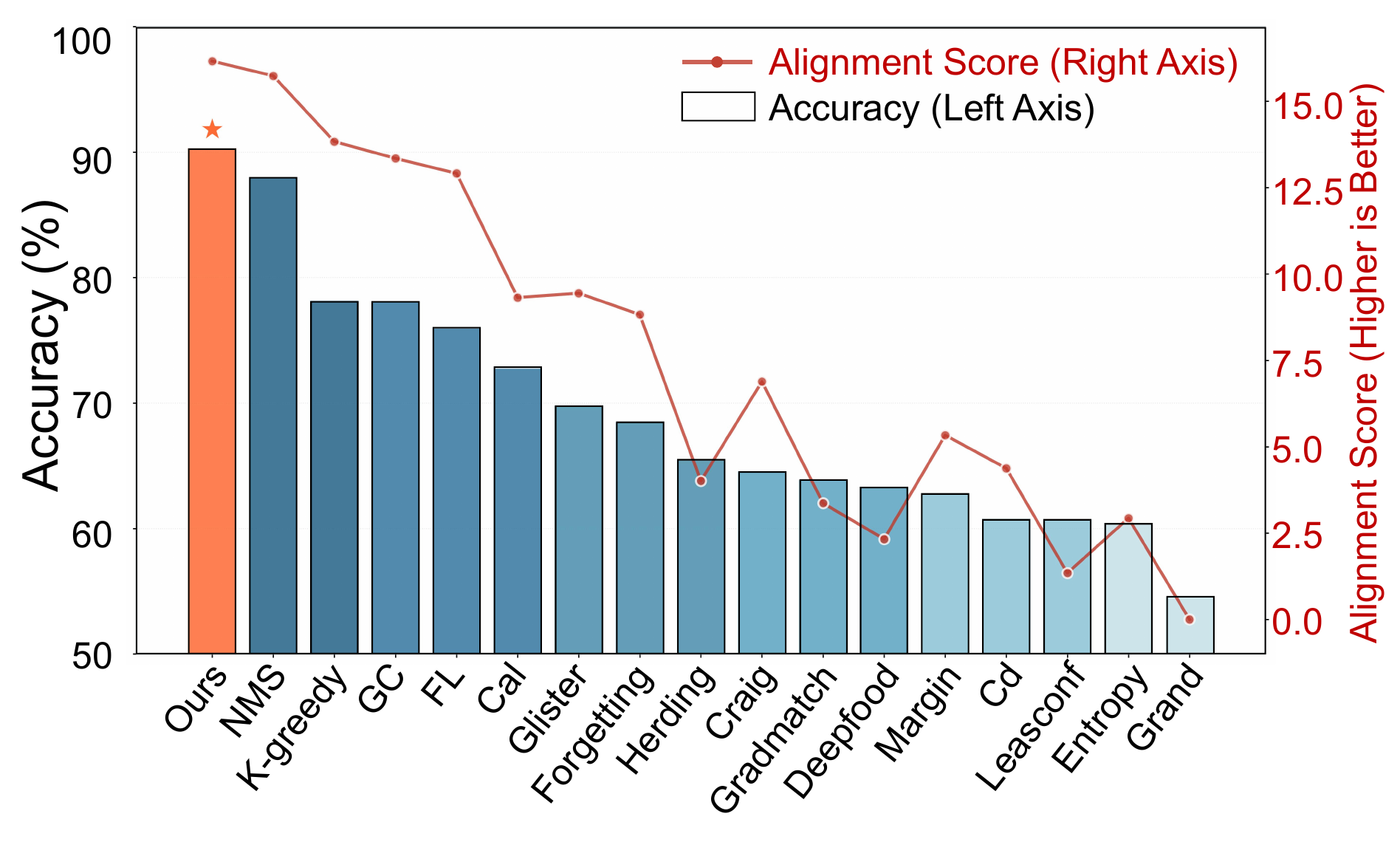}
    \vspace{-3mm}
    \caption{Relationship between downstream training accuracy and distributional equivalence. Results indicate that enforcing distributional equivalence leads to improved performance.} \vspace{-4mm}
    \label{fig:accuracy deviation}
\end{figure}

This SOTA performance derives from optimizing full distributional equivalence rather than model-dependent information and heuristics (e.g., \textit{learning difficulty} or \textit{training trajectory}). As shown in Fig.~\ref{fig:accuracy deviation}, we observe a positive correlation between the coreset's distributional alignment and downstream accuracy, demonstrating the robustness of this model-agnostic approach. The exceptional performance on complex datasets directly stems from the performance of PD-CFD to captures the higher-order moments and fine-grained structures (textures, edges) that other metrics miss while also resolving the \textit{vanishing phase gradient} problem. 

\begin{figure}
    \centering
    \includegraphics[width=0.9\linewidth]{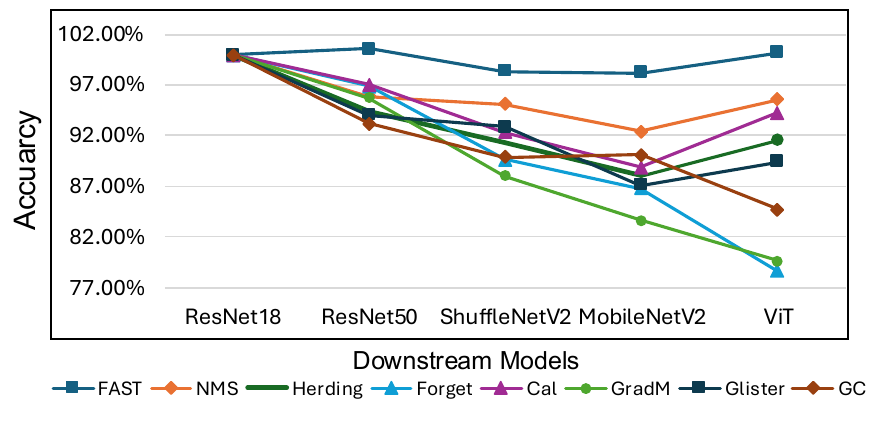}
    \vspace{-5mm}
    \caption{Cross-model generalization (Normalized). DNN-based methods suffer from performance degradation when transferring to other architectures while DNN-free approaches (FAST and NMS) remain stable performance.}
    \vspace{-5mm}
    \label{fig:generalization}
\end{figure}
\begin{figure}
    \centering
    \includegraphics[width=0.8\linewidth]{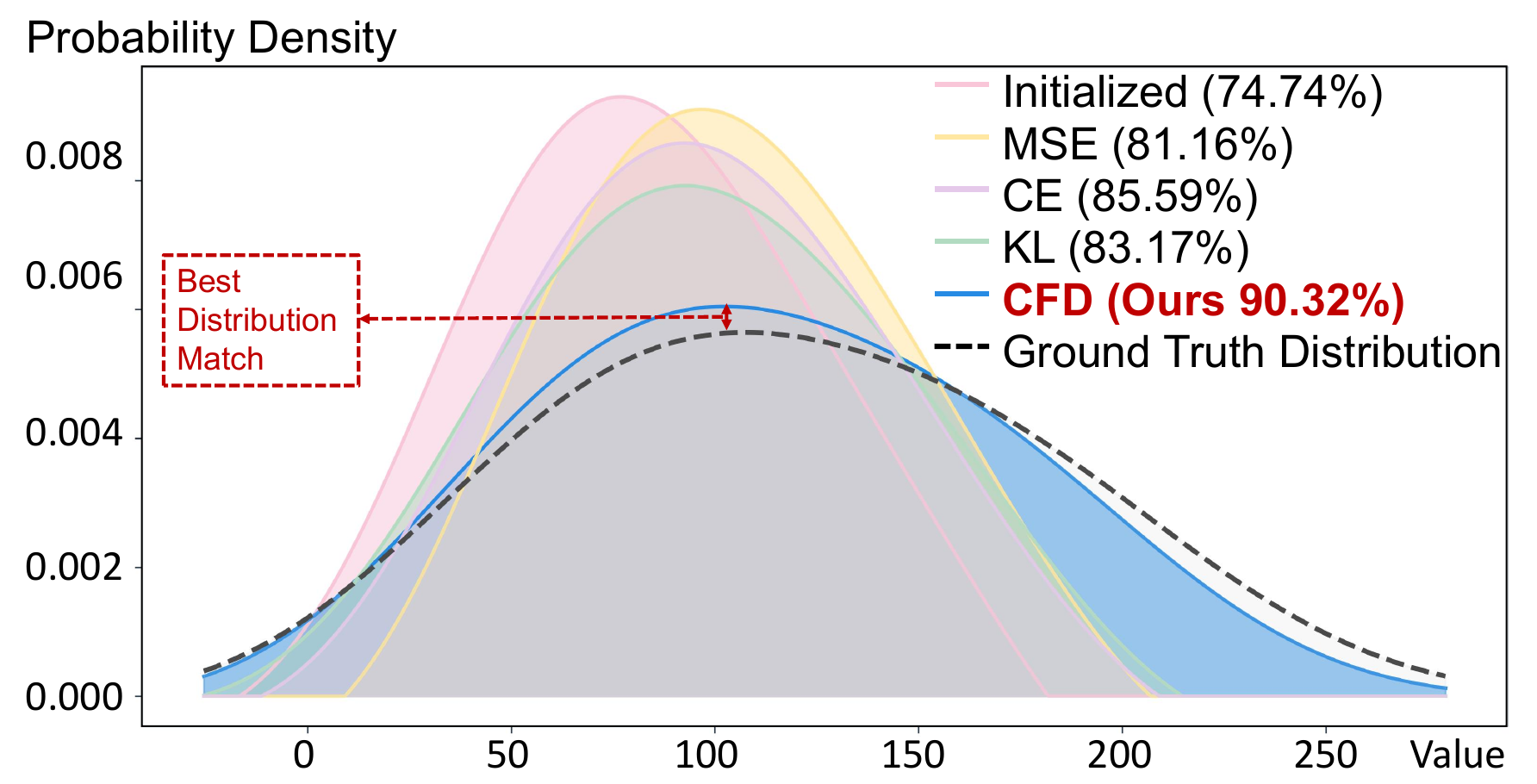}
    \vspace{-2mm}
    \caption{Ablation on distribution alignment with different metrics}
    \vspace{-6mm}
    \label{fig:metrics ablation}
\end{figure}
\begin{figure*}[t]
    \centering
    \includegraphics[width=0.9\linewidth]{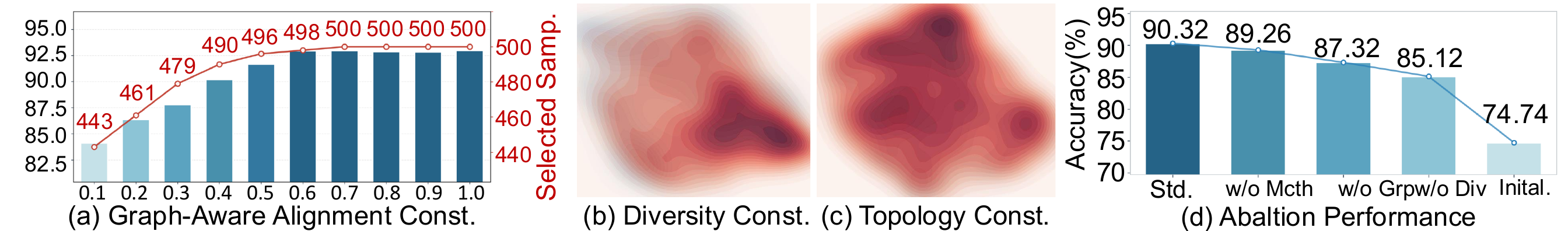} \vspace{-2mm}
    \caption{Ablation on graph regulations. Removing any of the three constraints—(a) GUNN ($\mathcal{L}_{align}$), (b) DPP ($\mathcal{L}_{div}$), or (c) graph regularization ($\mathcal{L}_{graph}$)—leads to degraded alignment, coverage, or topology, confirming their complementary roles in maintaining stability.} \vspace{-5mm}
    \label{fig:graph constrains}
\end{figure*}

\noindent\textbf{Cross-Architecture Generalization.} As demonstrated in Fig.~\ref{fig:generalization}, FAST exhibits strong cross-architecture generalization. When evaluated across diverse architectures at multiple keep rates {10\%, 20\%, 30\%}, our coreset shows a negligible accuracy drop (average $0.53$\%, even even achieves accuracy gains on some models) upon cross-architecture transfer, while competing DNN-free and DNN-based methods degrade by 3.02\% and 8.68\% on average.

This strong generalization confirms that FAST is a truly DNN-free method which seeks fundamental distributional equivalence rather than fitting a proxy model. By avoiding architectural bias, the resulting coreset faithfully mimics the full dataset’s behavior across diverse downstream scenarios, embodying a valuable \textit{Write once, run anywhere} property.

\noindent\textbf{Computational Efficiency.} Given the importance of computational efficiency, especially on resource-constrained edge devices, we evaluate FAST's efficiency on CIFAR-10 (10\% keep rate) against strong baselines. As detailed in Table~\ref{tab:energy}, we assess the total cost (sampling and training with early stopping) via runtime and energy. As shown, DNN-based methods, that require network inference for sampling, inevitably suffer from high energy consumption and slow sampling speeds. While gradient or trajectory-based methods may converge faster during training, their substantial sampling overhead negates this benefit. In contrast, FAST not only drastically reduces sampling time but also enables downstream models to achieve comparable accuracy with fewer training epochs. Furthermore, the \textit{FAST-edge} results in Table~\ref{tab:energy}, confirming its potential to be deployed on edge devices with limited resources.

\noindent\textbf{Performance on LLM Datasets.} We evaluate FAST on LLM tuning using the Alpaca \cite{alpaca2023} dataset. Coresets with 10\%, 20\% and 30\% keep ratios are sampled to finetune LLaMA-7B \cite{touvron_llama_2023}and the resulting models were assessed on the InstructEval benchmark \cite{instructeval2024}. FAST surpasses the SOTA DNN-free method by 2.6\% in average accuracy, demonstrating its effectiveness on semantically rich LLM tasks. Detailed results and discussion are provided in Appendix.

\vspace{-1mm}
\subsection{Ablation}
\vspace{-2mm}
\noindent\textbf{Ablation on Distribution Metrics.} Building on our analysis in Fig.~\ref{fig:alignment compare}, we ablate the distribution metric on CIFAR-10 (10\% keep rate, ResNet18 backbone) by replacing PD-CFD with KL, CE, and MSE, holding all other parameters constant. As shown in Fig.~\ref{fig:metrics ablation}, using MSE as the metric results in significant high-order moment deviations and a 9.16\% accuracy drop. Similarly, KL and CE, while improving variance alignment, still fail to match skewness and kurtosis, causing accuracy drops of 7.15\% and 4.73\%, respectively.

Since the effectiveness of the MMD method directly depends on the chosen kernel function, we will not expand the discussion here. But when the chosen kernel is insufficient to align high-order moments, the MMD method still cannot guarantee complete distribution alignment. We provide more detailed proof in Appendix.

\noindent\textbf{Ablation on Spectral Graph Regularization.}
We conduct an ablation study to quantify the efficacy of our graph constraints. Results in Fig.~\ref{fig:graph constrains} demonstrate the necessity of each component. Lacking GUNN ($\mathcal{L}_{align}$), optimized points \textit{clump} and map to a few identical samples, causing degradation in sampling quality. Without DPP ($\mathcal{L}_{div}$), points are attracted to a few high-density modes, resulting in poor coverage. Removing graph regularization ($\mathcal{L}_{graph}$) disrupts local topology by inducing excessive local clustering. 

These failures highlight the core difficulty of addressing a discrete sampling task through continuous optimization. The continuous-to-discrete gap causes gradient descent (with only $\mathcal{L}_{CF}$) to converge to degenerate solutions where multiple proxies $\tilde{y}_i$ collapse into a single mode, shrinking the effective coreset. Our constraints bridge this gap: DPP ($\mathcal{L}_{div}$) enforces spatial diversity in the continuous space, while GUNN ($\mathcal{L}_{align}$) and $\mathcal{L}_{graph}$ ensures topological consistency with the discrete manifold, preventing mode collapse and ensuring a stable, representative coreset.

\begin{figure}
    \centering
    \includegraphics[width=0.7\linewidth]{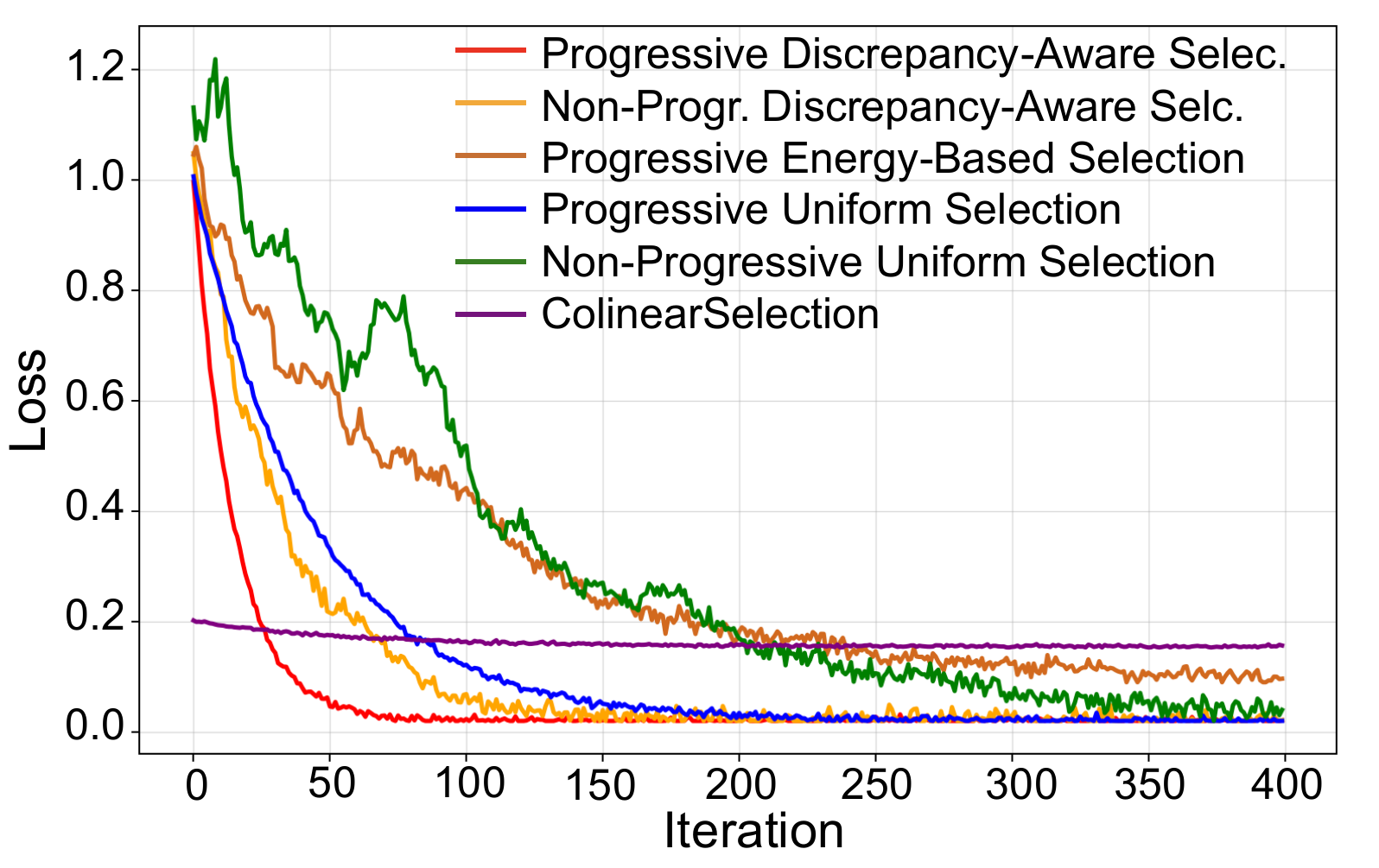}
    \vspace{-3mm}
    \caption{Ablation on PDAS. PDAS achieves stable and rapid convergence by progressively selecting discriminative frequencies, while other strategies yield unstable and suboptimal optimization.}
    \vspace{-5mm}
    \label{fig:converenge}
\end{figure}

\begin{figure}
    \centering
    \includegraphics[width=0.75\linewidth]{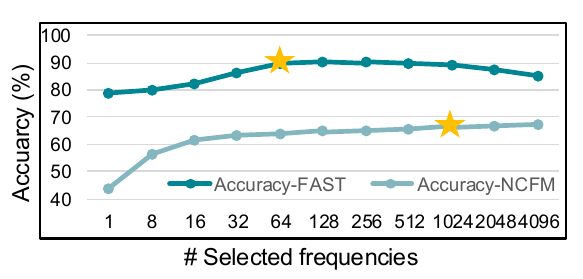}
    \vspace{-4mm}
    \caption{Effect of the number of selected frequencies. }
    \vspace{-6mm}
    \label{fig:freq num}
\end{figure}

\noindent\textbf{Ablation on Frequency Selection.} 
We validate our PDAS by fixing the frequency count (64) and comparing it against baselines in Fig.~\ref{fig:converenge}. The results confirm the critical importance of the selection strategy: (1) A worst-case collinear set degenerates the effectiveness of the CFD metric, causing optimization failure. (2) Magnitude-based Top-K selection performs poorly, as it ignores the distributional discrepancies and fails to select the most discriminative frequencies. (3) Non-progressive strategies are also suboptimal, as their erratic global selection (e.g., alternating between low and high frequencies) leads to unstable convergence. In contrast, our adaptive, curriculum-based PDAS achieves optimal matching, stably converges  within fewer iterations .

We further analyze the impact of the total frequency count, as illustrated in Fig.~\ref{fig:freq num}. FAST saturates significantly faster and requires far fewer frequencies to reach optimal performance than the NCFM baseline. This high frequency utilization efficiency is attributable to PDAS prioritizing frequencies with the largest contribution to the current distributional discrepancy under pre-optimized anisotropic frequency distribution. This not only yields superior performance but also reduces the computational overhead of CFD.

\begin{figure}
    \centering
    \includegraphics[width=0.75\linewidth]{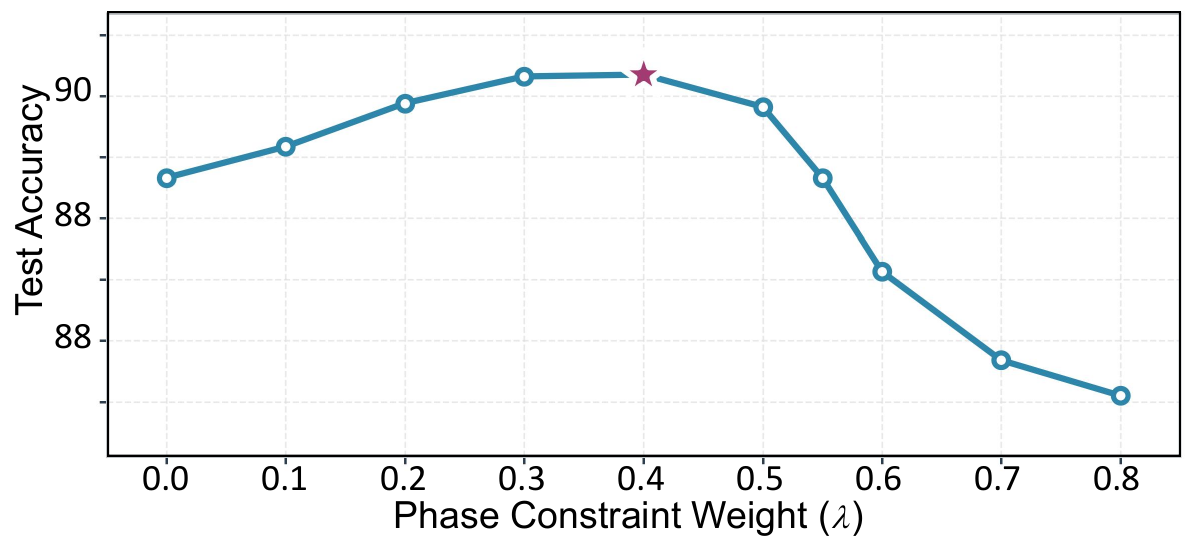}
    \vspace{-3mm}
    \caption{Ablation on phase-decoupled constraint. Performance peaks at $\lambda_{p}=0.4$, while larger weights introduce noise and degrade alignment.}
    \vspace{-8.5mm}
    \label{fig:phase ablation}
\end{figure}

\noindent\textbf{Ablation on Phase Constraint.}
We evaluate the phase-decoupled constraint ($\lambda_{p}$) on the detail-rich RESISC45 dataset ($\alpha$ fixed at $1.2$). As shown in Fig.~\ref{fig:phase ablation}, performance peaks at $\lambda_{p}=0.3$, confirming its effectiveness in reinforcing high-frequency structures. For $\lambda_{p}>0.4$, overemphasis on phase leads to high-frequency noise and degraded global consistency, reducing overall distributional alignment. To verify its generality, we integrate the optimal $\lambda_{p}=0.3$ into NCFM \cite{wang25ncfm}, achieving a 19.12\% accuracy gain on the detail-rich CUB-200-2011 \cite{wah11cub200} dataset at IPC=10, demonstrating its robust ability to capture fine-grained structures. 
Detailed results are provided in the Appendix.

\section{Conclusion}
In this work, we introduced FAST, the first DNN-free distribution-matching coreset selection framework that formulates coreset selection task as a graph-constrained optimization problem grounded in spectral graph theory and employs the Characteristic Function Distance (CFD) to capture full distributional information (i.e., all moments and intrinsic correlations) in the frequency domain. Our phase-decoupled CFD formulation further resolves the amplitude–phase coupling that obscures medium and high-frequency structural information, enabling accurate preservation of fine-grained semantics and high-order dependencies.

Extensive experiments across image classification, fine-grained recognition, texture and remote-sensing datasets, and LLM instruction tuning demonstrate that FAST consistently outperforms both DNN-based and DNN-free baselines. Moreover, FAST's excellent performance on LLM datasets further suggests that high-level semantic structure can be retained through spectral-graph-based distribution alignment, without explicit reliance on neural feature extractors. By leveraging geometric and frequency-domain signals, FAST preserves the underlying semantic neighborhood relations required for downstream reasoning. These experiments jointly demonstrate the robustness and broad generalization ability of FAST, even on tasks that rely heavily on semantic consistency.

Beyond accuracy, FAST also delivers significant efficiency gains. Since FAST operates completely without neural network inference, and our PDAS strategy prioritizes only the most discriminative frequencies, the computation required for distribution alignment is drastically reduced. Furthermore, our graph-based manifold construction avoids the repeated forward passes central to traditional gradient- or loss-based methods, which significantly lowers the sampling cost and highlights FAST's practical utility for resource-constrained environments.

{
    \small
    \bibliographystyle{ieeenat_fullname}
    \bibliography{main}
}
\newpage
\appendix



\begin{center}
\Large
\vspace{0.5em}Supplementary Material \\
\vspace{1.0em}
\end{center}

\newtheorem{definition}{Definition}
\newtheorem{proposition}{Proposition}
\newtheorem{theorem}{Theorem}
\newtheorem{lemma}{Lemma}
\newtheorem{corollary}{Corollary}
\newtheorem{example}{Example}
\newtheorem{remark}{Remark}

\section{Computational Overhead}

\vspace{-5mm}

\begin{table}[htbp]
\centering
\footnotesize
\setlength{\tabcolsep}{2pt}
\renewcommand{\arraystretch}{1.1}
\begin{tabular}{@{}lccr@{}}
\toprule
Module/Eq. & Complexity/$O(\cdot)$ & Total FLOPs & Peak RAM \\
\midrule
Graph Construction (1) & $N_c^2 \cdot D$ & $7.41 \times 10^{14}$ & $\mathbf{778\,\text{MB}}$ \\
Spectral Embedding (2) & $N_c^3$ & $2.10 \times 10^{12}$ & $163\,\text{KB}$ \\
Cost Matrix (3) & $M_c \cdot N_c \cdot d$ & $6.30 \times 10^{12}$ & $655\,\text{KB}$ \\
Hungarian Match& $M_c \cdot N_c^2$ & $8.40 \times 10^{13}$ & $741\,\text{KB}$  \\
PD-CFD Loss (9) & $(N_c+M_c) N_{freq} d$ & $9.20 \times 10^{12}$ & $1.03\,\text{MB}$  \\
\midrule
\textbf{Total FAST}  & -- & $8.43 \times 10^{14}$ & $\approx 782\,\text{MB}$  \\
ResNet-50 Comparison & -- & $5.25 \times 10^{15}$ & $\approx 6.25 \,\text{GB}$  \\
\bottomrule
\end{tabular}
\medskip
\vspace{-4mm}
\caption{Expensive Computation Breakdown on ImageNet-1K.}

\noindent\footnotesize
\textbf{Note:}
FAST assumes per-class selection. ResNet-50 FLOPs are for one epoch inference on ImageNet-1K with a batch size of 256. Total feature extraction overhead of FAST on ImageNet-1K is 16\% of  ResNet-50 .
\vspace{-6mm}
\end{table}

\begin{table}[h]
\centering
\resizebox{1.0\linewidth}{!}{
\begin{tabular}{lccccccc}
\toprule
\textbf{Keep Rate} & \textbf{10\%} & \textbf{Time(h)} & \textbf{20\%} & \textbf{Time(h)} & \textbf{30\%} & \textbf{Time(h)} & \textbf{Full} \\
\midrule
previous SOTA$^\ddagger$ & 53.6 & 7.9 & 62.6 & 21.1 & 64.5 & 30.2 & 71.2 \\
FAST$^\ddagger$ & \textbf{56.2} & \textbf{2.5} & \textbf{63.9} & \textbf{6.1} & \textbf{66.9} & \textbf{8.2} & 71.2 \\
\bottomrule
\end{tabular}
}
\vspace{-2mm}
\caption{Performance Comparison on ImageNet-1K. $^\ddagger$On CPU.}
\vspace{-6mm}
\end{table}

\vspace{-1mm}

\section{Optimization Stability}
\vspace{-2mm}

The \textit{Hungarian mapping} evolves from dynamic exploration to stable anchoring, ensuring robust continuous-to-discrete alignment. Despite discrete updates to $\pi$, the loss maintains continuity at cost-equalizing \textit{Voronoi boundaries}, and divergence is prevented by topological constraints that limit gradient fluctuations to the local manifold neighborhood. Since \textit{Hungarian algorithm} minimizes the transport cost globally, gradient descent on the smooth objective $\tilde{Y}$ ensures $\mathcal{L}(\tilde{Y}^{(t+1)}, \pi^{(t+1)}) \le \mathcal{L}(\tilde{Y}^{(t)}, \pi^{(t)})$, yielding monotonic descent optimization: $0 \le \mathcal{L}_{total}^{(t+1)} \le \mathcal{L}_{total}^{(t)}$, where \textit{Monotone Convergence Theorem} guarantees convergence.

\section{Necessity of Embeddings}
\vspace{-1.5mm}

In LLM experiments, we utilize general Sentence-BERT embeddings as standard preprocessing for manifold construction. The structural distinction between the encoder-only embedder and decoder-only LLM introduces a general semantic prior rather than coupled architectural bias, confirmed by Table \ref{tab:llm-exps} that shows FAST's robust generalization.

\vspace{-2mm}

\begin{table}[h]
\centering
\footnotesize
\setlength{\tabcolsep}{3pt}
\renewcommand{\arraystretch}{1.1}
\resizebox{\linewidth}{!}{
\begin{tabular}{lcccccc}
\toprule
\textbf{Model} & \textbf{L-FAST} & \textbf{L-Rand} & \textbf{Q-FAST} & \textbf{Q-Rand} & \textbf{M-FAST} & \textbf{M-Rand} \\
\midrule
Acc (\%) & $39.0_{\textcolor{red}{\uparrow\,8.7}}$ & $35.9_{\uparrow\,5.6}$ & 
$72.3_{\textcolor{red}{\uparrow\,6.1}}$ & 
$70.9_{\uparrow\,4.7}$ & 
$63.1_{\textcolor{red}{\uparrow\,4.0}}$ & 
$61.6_{\uparrow\,2.5}$ \\
\bottomrule
\end{tabular}
}
\noindent\footnotesize
\textbf{Notation:}
L: LLaMA2-7B, Q: Qwen2.5-7B, M: Mistral-7B, Rand:Random. $\uparrow$ indicates the relative improvement to the base model. 
\vspace{-2mm}
\caption{Supplementary Experiments on LLM datasets.}
\label{tab:llm-exps}
\end{table}
\vspace{-4mm}
\section{More Ablation Studies}

Use CIFAR-10 dataset (10\% KR, training ResNet-18) unless noted; random sampling shows 75.70\%\textsubscript{$\pm$7.63} accuracy.

\begin{table}[h]
\centering
\resizebox{0.95\linewidth}{!}{
\begin{tabular}{lccccc}
\toprule
\textbf{Method} & \textbf{Acc (\%)} & \textbf{Time (s)} & \textbf{Energy (Wh)} & \textbf{Std} & \textbf{Opt/Steps} \\
\midrule
Pixel-Opt$^\ddagger$ & 61.52 & 3770 & 29.75 & 13.73 & 1300 \\
ResNet-50$^\dagger$ & 81.51 & 627.67 & 37.75 & 3.32 & 230 \\
FAST$^\ddagger$ & \textbf{90.32} & \textbf{353} & \textbf{1.409} & \textbf{1.21} & \textbf{80} \\

\bottomrule
\end{tabular}
}
\vspace{-2mm}

\caption{Graph Feature Extractor Ablation on $^\ddagger$CPU, $^\dagger$GPU.}
\vspace{-4mm}
\end{table}

\noindent \textbf{Analysis:} Pixel-level optimization fails due to the sparsity and noise of high-dimensional space, while incurring massive computational costs ($10\times$ slower). ResNet-50 features suffer from architectural bias and computational overhead.
\vspace{-2mm}

\begin{table}[h]
\centering
\resizebox{\linewidth}{!}{
\begin{tabular}{l|c|cc|cc}
\toprule
\textbf{Parameter} & \textbf{Ours} & \boldmath{$k=5$} & \boldmath{$k=50$} & \boldmath{$d=8$} & \boldmath{$d=128$} \\
\midrule
Trustworthiness & \textbf{0.95\textsubscript{$\pm$0.01}} & 0.73\textsubscript{$\pm$0.10} & 0.52\textsubscript{$\pm$0.09} & 0.44\textsubscript{$\pm$0.09} & 0.92\textsubscript{$\pm$0.02} \\
Continuity      & 0.91\textsubscript{$\pm$0.03}          & 0.49\textsubscript{$\pm$0.15} & 0.93\textsubscript{$\pm$0.06} & 0.92\textsubscript{$\pm$0.03} & 0.87\textsubscript{$\pm$0.06} \\
Acc (\%)        & \textbf{90.32\textsubscript{$\pm$1.21}} & 72.12\textsubscript{$\pm$5.56} & 85.01\textsubscript{$\pm$3.27} & 78.86\textsubscript{$\pm$3.13} & 86.73\textsubscript{$\pm$4.16}$^*$ \\
\bottomrule
\end{tabular}
}
\vspace{-2mm}
\caption{Hyperparameter Ablation (neighbor scale $k$ and reduced dimension $d$). $^*$Time doubles. FAST adopts $k=15$ and $d=32$.}
\end{table}

\noindent \textbf{Analysis:} Small $k$ fractures the manifold, while large $k$ causes collapse, connecting distant classes; low $d$ bottlenecks information, and high $d$ adds noise and doubles optimization time. Our settings optimally preserve topology.

\begin{table}[h]
\centering
\setlength{\tabcolsep}{1.5pt}
\renewcommand{\arraystretch}{1.1}
\resizebox{1.0\linewidth}{!}{
\begin{tabular}{lcccccc}
\toprule
\textbf{Strategy} & \textbf{DAS} & \textbf{PES} & \textbf{PUS} & \textbf{US} & \textbf{Collinear} & \textbf{PDAS} \\
\midrule
Acc (\%) & 
88.89\textsubscript{$\pm$2.23} & 
87.12\textsubscript{$\pm$2.60} & 
88.66\textsubscript{$\pm$2.01} & 
86.60\textsubscript{$\pm$2.73} & 
74.16\textsubscript{$\pm$7.77} & 
\textbf{90.32\textsubscript{$\pm$1.21}} \\
Opt/Steps & 150 & 400 & 250 & 430 & 100$^*$ & \textbf{80} \\
\bottomrule
\end{tabular}
}
\vspace{-2mm}
\caption{Impact of Frequency Selection. $^*$Optimization failure mode. Illustration and abbreviations, please refer to paper Fig. 9.}
\end{table}

\noindent \textbf{Analysis:} Baselines unstably converge due to suboptimal frequency selection; PDAS achieves optimal stability by progressively selecting discriminative frequencies.

\vspace{-2mm}

\begin{table}[h]
\centering
\small
\resizebox{0.8\linewidth}{!}{
\begin{tabular}{lccc}
\toprule
\textbf{Method} & \textbf{CIFAR-10} & \textbf{RESISC-45} & \textbf{DTD} \\
\midrule
FAST-Vanilla & 88.17\textsubscript{$\pm$2.12} & 82.01\textsubscript{$\pm$3.99} & 41.15\textsubscript{$\pm$3.20} \\
FAST-PD  & \textbf{90.32\textsubscript{$\pm$1.21}} & \textbf{85.00\textsubscript{$\pm$2.01}} & \textbf{45.77\textsubscript{$\pm$2.53}} \\
\bottomrule
\end{tabular}
}
\vspace{-2mm}
\caption{Phase-Decoupled CFD Ablation (full ablation in Fig. 7).}
\end{table}

\noindent \textbf{Analysis:} \textit{CFD} fails to capture high-frequency details due to amplitude-phase coupling. \textit{PD-CFD} resolves this, showing substantial gains on texture-rich DTD (+4.62\%) and RESISC-45 (+2.99\%) compared to CIFAR-10 (+2.15\%). 

\section{Additional Analysis on Limitations of KL and CE for High-Order Moment Alignment}
\label{sec:appendixa}

\paragraph{Notation.}
Let $P$ be an unknown target distribution on $\mathcal{X}\subseteq\mathbb{R}^d$ with density $p$.
An exponential family is
\[
\mathcal{Q}_T
~:=~
\Big\{ q_\theta(x)=h(x)\exp\!\big(\theta^\top T(x)-A(\theta)\big) \;\Big|\; \theta\in\Theta\subset\mathbb{R}^m \Big\},
\]
where $T:\mathcal{X}\to\mathbb{R}^m$ is a vector of sufficient statistics,
$A(\theta):=\log\int h(x)\exp(\theta^\top T(x))\,dx$ is the log-partition function (cumulant generating function),
and $h$ is a base density.
We write $\mu(\theta):=\nabla A(\theta)=\mathbb{E}_{q_\theta}[T(X)]$ for the mean-parameter map.

For $k\in\mathbb{N}$, define the (truncated) polynomial statistics
\[
T^{(\le k)}(x) \;=\; \big( x,~ xx^\top,~ x^{\otimes 3},~\dots,~x^{\otimes k}\big),
\]
so that $\mathcal{Q}_{\le k}$ denotes the exponential family whose sufficient statistics are all monomials up to degree $k$
(with a suitable choice of $h$ to ensure normalizability).

\vspace{0.5em}
\begin{proposition}[I-projection yields moment matching on the chosen statistics]
\label{prop:iprojection-moment-matching}
Consider the KL minimization (I-projection) of $P$ onto an exponential family $\mathcal{Q}_T$:
\begin{equation}
\begin{split}
\theta^\star \;&\in\; \arg\min_{\theta\in\Theta}\; D_{\mathrm{KL}}\!\big(P \,\Vert\, Q_\theta\big) \\
&~=~ \arg\min_{\theta}~ \int p(x)\log\frac{p(x)}{q_\theta(x)}\,dx.
\end{split}
\end{equation}
Then $A$ is convex and the KL objective is strictly convex in $\theta$ on the mean-parameter interior; the unique minimizer
$\theta^\star$ (when it exists) satisfies the first-order optimality condition
\[
\nabla_\theta\!\Big(\mathbb{E}_{P}[\theta^\top T(X)]-A(\theta)\Big)\Big|_{\theta=\theta^\star}
~=~ \mathbb{E}_{P}[T(X)] - \nabla A(\theta^\star)~=~0.
\]
Equivalently,
\[
\;\mathbb{E}_{P}[T(X)] ~=~ \mathbb{E}_{Q_{\theta^\star}}[T(X)]\;.
\]
\end{proposition}

\begin{proof}[Proof (via convex duality and Fenchel--Legendre conjugacy)]
Expanding the KL divergence,
\[
D_{\mathrm{KL}}(P\Vert Q_\theta)
= -H(P) - \mathbb{E}_{P}\!\big[\theta^\top T(X)-A(\theta)+\log h(X)\big],
\]
where $H(P)$ and $\mathbb{E}_{P}[\log h(X)]$ are $\theta$-independent constants.
Thus minimizing $D_{\mathrm{KL}}(P\Vert Q_\theta)$ is equivalent to maximizing the concave functional
\[
\mathcal{L}(\theta) := \mathbb{E}_{P}[\theta^\top T(X)] - A(\theta).
\]
Since $A$ is convex (indeed, $A$ is the log-moment generating function of $T$ under $h$), $\mathcal{L}$ is concave and (under standard interiority conditions) has a unique maximizer $\theta^\star$.
The Fenchel--Legendre conjugate of $A$ is $A^\ast(\mu):=\sup_{\theta}\{\langle\theta,\mu\rangle - A(\theta)\}$; strong duality yields that the optimal $\theta^\star$ satisfies
$\nabla A(\theta^\star)=\mu^\star=\mathbb{E}_{P}[T(X)]$ (i.e.\ the moment-matching equations).
Finally $\nabla A(\theta)=\mathbb{E}_{Q_\theta}[T(X)]$ by standard exponential-family calculus, completing the proof.
\end{proof}

\vspace{0.5em}
\begin{corollary}[Distributions controlled by KL under restricted sufficient statistics]
\label{cor:restricted}
Let $\mathcal{Q}_T$ be any exponential family with sufficient statistics $T$.
Then the I-projection $Q_{\theta^\star}$ aligns exactly those coordinates of $T$:
\[
\mathbb{E}_{P}[T_i(X)] = \mathbb{E}_{Q_{\theta^\star}}[T_i(X)] \quad \text{for each component } T_i,
\]
but imposes no necessary constraint on expectations of functions $f$ lying outside the linear span of $\{1, T_1,\dots,T_m\}$.
In particular, if $T=T^{(\le k)}$ contains all monomials up to degree $k$, then KL minimization guarantees matching of all raw moments up to order $k$, and \emph{does not} in general constrain any $(k\!+\!1)$-st or higher-order moments.
\end{corollary}

\begin{proof}
Immediate from Proposition~\ref{prop:iprojection-moment-matching}, noting that the optimality system is exactly the linear system matching $\mathbb{E}_{P}[T]$ and $\mathbb{E}_{Q_{\theta}}[T]$.
Any $f$ outside the closed linear span of $\{1,T\}$ cannot be represented as $\alpha_0 + \alpha^\top T$, hence its expectation is unconstrained by the KKT system.
\end{proof}

\vspace{0.5em}
\begin{corollary}[Gaussian family ($k=2$) controls mean and covariance but not higher cumulants]
\label{cor:gaussian}
Let $\mathcal{Q}_{\mathcal{N}}=\{\mathcal{N}(\mu,\Sigma)\}$.
Then $\mathcal{Q}_{\mathcal{N}}$ is an exponential family with
$T(x)=(x,\,xx^\top)$ and
$\nabla A(\mu,\Sigma) = (\mathbb{E}[X],\,\mathbb{E}[XX^\top])$.
The I-projection $Q_{\theta^\star}=\mathcal{N}(\mu^\star,\Sigma^\star)$ thus satisfies
\[
\mathbb{E}_{Q_{\theta^\star}}[X] = \mathbb{E}_{P}[X],
\qquad
\mathbb{E}_{Q_{\theta^\star}}[XX^\top] = \mathbb{E}_{P}[XX^\top],
\]
but neither skewness (third cumulant) nor kurtosis (fourth cumulant) is constrained or minimized by the KL objective in general.
\end{corollary}

\begin{proof}
A direct specialization of Corollary~\ref{cor:restricted} with $k=2$.
\end{proof}

\vspace{0.5em}
\begin{proposition}[Augmenting statistics raises the matched moment order, but only up to that order]
\label{prop:ladder}
Let $\mathcal{Q}_{\le k}$ denote the exponential family with $T^{(\le k)}$.
Then the sequence of I-projections $\{Q^{(k)}\}_{k\ge 1}$ defined by
\[
Q^{(k)} \in \arg\min_{Q\in\mathcal{Q}_{\le k}} D_{\mathrm{KL}}(P\Vert Q)
\]
satisfies, for each fixed $k$,
\[
\mathbb{E}_{Q^{(k)}}[X^{\otimes r}] ~=~ \mathbb{E}_{P}[X^{\otimes r}]
\quad\text{for all } r=1,2,\dots,k,
\]
and there is no general guarantee that $\mathbb{E}_{Q^{(k)}}[X^{\otimes r}]$ aligns with $\mathbb{E}_{P}[X^{\otimes r}]$ for any $r>k$.
\end{proposition}

\begin{proof}
By Proposition~\ref{prop:iprojection-moment-matching} applied to $T^{(\le k)}$, the optimality conditions enforce equality of the first $k$ raw-moment tensors. Since the KL objective reduces to a linear functional of $T$ minus $A(\theta)$, any statistics not included in $T$ do not appear in the optimality system and remain uncontrolled.
\end{proof}

\vspace{0.5em}
\begin{remark}[Consequences for KL/CE-based \emph{evaluation} of distributional alignment]
\label{rem:evaluation}
Since cross-entropy minimization is equivalent to minimizing $D_{\mathrm{KL}}(P\Vert Q_\theta)$ when $P$ is fixed,
Propositions~\ref{prop:iprojection-moment-matching}--\ref{prop:ladder} imply a fundamental limitation:
\emph{Under a restricted exponential family, KL/CE can certify alignment of at most the moments encoded in $T$ (e.g., mean and covariance for Gaussians), and it provides no guarantees for higher-order moments or cumulants that are not present in $T$.}
Hence, unless higher-order statistics are explicitly included in the model (i.e., increasing $k$), KL/CE-based fitting and evaluation do not ensure  moment alignment.
\end{remark}

\vspace{0.5em}
\begin{example}[Gaussian vs.\ heavy-tailed target: kurtosis mismatch at the KL optimum]
\label{ex:gaussian-laplace}
Let $P$ be a zero-mean heavy-tailed distribution (e.g., a centered Laplace in 1D) with variance $\sigma^2$ and kurtosis $\kappa_P>3$.
Its I-projection onto $\mathcal{Q}_{\mathcal{N}}$ is $Q^\star=\mathcal{N}(0,\sigma^2)$, which matches mean and variance by Corollary~\ref{cor:gaussian}.
However, $\kappa_{Q^\star}=3\neq \kappa_P$, i.e., fourth-order moments are not aligned at the KL optimum---a concrete manifestation of Remark~\ref{rem:evaluation}.
\end{example}

\vspace{0.5em}
\begin{remark}[Mixtures vs.\ single exponential families]
A Gaussian mixture model (GMM) is a convex combination of Gaussians and \emph{not} a single exponential family with a fixed finite-dimensional $T$.
Therefore, the structural limitation in Propositions~\ref{prop:iprojection-moment-matching}--\ref{prop:ladder} does not apply in the same form to GMMs: with sufficiently many components, a GMM can approximate higher-order structures arbitrarily well.
However, practical optimization remains nonconvex and capacity-limited, so  alignment may still fail in practice despite the absence of a finite-dimensional $T$.
\end{remark}

\vspace{0.6em}
\subsection*{A General Counterexample: Small KL yet Divergent Higher-Order Moments}

We now give a short, self-contained construction showing that (beyond model restrictions) \emph{the KL divergence itself does not control unbounded test-function expectations}, such as  moments.

\begin{proposition}[KL can be arbitrarily small while $k$-th moments diverge]\label{prop:counterexample}
Fix any integer $k\ge 3$. There exists a sequence of distributions $P_M$ and a reference $Q$ such that
\[
D_{\mathrm{KL}}(P_M\Vert Q)\to 0 \quad \text{but}\quad
\big|\mathbb E_{P_M}[X^k]-\mathbb E_{Q}[X^k]\big|\to\infty.
\]
\end{proposition}

\begin{proof}
Let $Q=\mathcal N(0,1)$ and $R_M=\mathcal N(M,1)$. Define the mixture
\[
P_M=(1-\varepsilon_M)\,Q+\varepsilon_M\,R_M,\qquad
\varepsilon_M=\frac{1}{M^{2}\log M}.
\]
\emph{(i) KL upper bound by joint convexity.}
By convexity of $D_{\mathrm{KL}}(\cdot\Vert Q)$ in its first argument,
\begin{equation}
\begin{split}
D_{\mathrm{KL}}&\!\big(P_M\Vert Q\big)\le (1-\varepsilon_M)D_{\mathrm{KL}}(Q\Vert Q)+\varepsilon_M D_{\mathrm{KL}}(R_M\Vert Q) \\
&=\varepsilon_M \cdot \tfrac{M^2}{2}=\tfrac{1}{2\log M}\to 0,
\end{split}
\end{equation}

where $D_{\mathrm{KL}}(\mathcal N(M,1)\Vert \mathcal N(0,1))=\tfrac{M^2}{2}$.

\noindent\emph{(ii) Divergence of the $k$-th moment gap.}
For $k\ge 3$, $\mathbb E_{R_M}[X^k]\sim M^k$. Hence
\begin{equation}
\begin{split}
&\big|\mathbb E_{P_M}[X^k]-\mathbb E_Q[X^k]\big|
= \varepsilon_M\big|\mathbb E_{R_M}[X^k]-\mathbb E_Q[X^k]\big| \\
&\gtrsim \varepsilon_M\,M^k
= \frac{M^{k-2}}{\log M}\xrightarrow[M\to\infty]{}\infty.
\end{split}
\end{equation}
Thus KL can be made arbitrarily small while the $k$-th moment discrepancy diverges.
\end{proof}

\paragraph{Interpretation.}
Pinsker’s inequality controls only \emph{bounded} test functions via total variation; polynomial moments are unbounded, so small KL does not imply closeness of higher moments. Proposition~\ref{prop:counterexample} complements the structural limitation (Propositions~\ref{prop:iprojection-moment-matching}--\ref{prop:ladder}) by showing a \emph{metric} limitation: even without any modeling restrictions, KL does not bound higher-order moments.

\section{Limitations of MMD for Aligning Higher-Order Statistics}
\label{appendix:mmd_limitation}

Maximum Mean Discrepancy (MMD) is widely used as a nonparametric
distributional metric, defined for a positive definite kernel $k$ by
\begin{equation}
\begin{split}
&\mathrm{MMD}_k(P,Q)
= \left\| \mu_P - \mu_Q \right\|_{\mathcal H_k} \\
&= \sup_{\|f\|_{\mathcal H_k}\le 1}
\left( \mathbb E_P[f(X)] - \mathbb E_Q[f(X)] \right),
\end{split}
\end{equation}
where $\mathcal H_k$ is the reproducing kernel Hilbert space (RKHS) associated
with $k$. When $k$ is characteristic, MMD metrizes weak convergence.
However, for the purpose of coreset selection—where the objective is to align
\emph{all} statistical moments (including heavy-tailed or high-order cumulants)—
weak convergence is insufficient. We show below that with commonly used bounded
kernels (e.g., Gaussian, Laplacian), MMD does \emph{not} control higher-order
moments, even when $\mathrm{MMD}_k(P,Q)$ is arbitrarily small.

\paragraph{Setup.}
Let $k(x,y)$ be any bounded positive definite kernel with
$\sup_{x,y}|k(x,y)| \le K < \infty$ (this includes the Gaussian RBF).
Let $Q$ be any reference distribution with finite moments of all orders.

We construct the following sequence:
\[
P_M = (1-\varepsilon_M) Q + \varepsilon_M R_M,
\]
where $R_M$ is a distribution concentrated at radius $\|x\|\approx M$, and
$\varepsilon_M$ is a vanishing mixing weight.

The construction is analogous to the KL example in
Proposition~\ref{prop:counterexample} but adapted to the MMD geometry.

\begin{proposition}[Small MMD does not control higher-order moments]
\label{prop:mmd_moment_gap}
For any bounded kernel $k$ and any integer $r\ge 3$, there exists a sequence
$P_M$ such that
\[
\mathrm{MMD}_k(P_M,Q) \longrightarrow 0 \quad\text{but}\quad
\big|\mathbb E_{P_M}[X^r] - \mathbb E_Q[X^r]\big|
\longrightarrow \infty.
\]
\end{proposition}

\begin{proof}
Because $k$ is bounded, the RKHS norm of the kernel mean embedding satisfies
\[
\|\mu_{P_M} - \mu_Q\|_{\mathcal H_k}
\le \mathbb E_{P_M,Q}[\,|k(X,Y)|\,] \le K.
\]
More precisely,
\[
\mu_{P_M}
= (1-\varepsilon_M)\mu_Q + \varepsilon_M\mu_{R_M},
\]
hence
\[
\mathrm{MMD}_k(P_M,Q)
= \varepsilon_M \|\mu_{R_M} - \mu_Q\|_{\mathcal H_k}
\le 2K\,\varepsilon_M.
\]
Taking $\varepsilon_M = 1/\log M$ yields
\[
\mathrm{MMD}_k(P_M,Q) \le \frac{2K}{\log M} \xrightarrow[M\to\infty]{} 0.
\]

Next, choose $R_M$ such that almost all its mass lies on $\|x\|\approx M$.
Then the $r$-th moment satisfies
\[
\mathbb E_{R_M}[X^r] \asymp M^r,
\]
and therefore
\begin{equation}
\begin{split}
\mathbb E_{P_M}[X^r]
&= (1-\varepsilon_M)\mathbb E_Q[X^r] + \varepsilon_M\,\mathbb E_{R_M}[X^r] \\
&\asymp \mathbb E_Q[X^r] + \varepsilon_M M^r.
\end{split}
\end{equation}
Since $\varepsilon_M M^r = M^r/\log M \to \infty$ for any $r\ge 3$,
we obtain
\[
\big|\mathbb E_{P_M}[X^r] - \mathbb E_Q[X^r]\big|
\longrightarrow \infty,
\]
even while $\mathrm{MMD}_k(P_M,Q)\to 0$.
\end{proof}

\paragraph{Interpretation.}
This result parallels Proposition~3 for KL: both reveal a \emph{metric
limitation} independent of modeling assumptions. For bounded kernels,
MMD metrizes weak convergence but cannot control expectations of
unbounded test functions—particularly polynomial functions that capture
higher-order moments, tail behavior, or heavy-tailed anisotropy.

Consequently, when MMD is used as an objective for coreset construction, moment alignment depends critically on the expressiveness of the chosen kernel. If the kernel is insufficient to probe heavy tails or high-order interactions,
the resulting coreset may perfectly match the MMD score yet deviate drastically in higher-order statistics—precisely the regime where our PD-CFD metric provides a more faithful discrepancy measure.

\section{Characteristic Functions: Complete Statistical Representation and Fourier–Analytic Foundations}

\paragraph{Notation.}
Let $P$ be a Borel probability measure on $\mathbb{R}^d$ with random vector $X\!\sim\!P$.
The \emph{characteristic function} (CF) of $P$ is
\[
\varphi_P(\omega)
~:=~
\int_{\mathbb{R}^d} e^{\,i\,\omega^\top x}\,P(dx)
~=~
\mathbb{E}_P[e^{\,i\,\omega^\top X}],\qquad \omega\in\mathbb{R}^d.
\]
Each $\varphi_P$ is bounded ($|\varphi_P|\le1$), uniformly continuous, and positive definite.

\vspace{0.5em}
\begin{proposition}[Fourier inversion: recovering $P$ from $\varphi_P$]
\label{prop:inversion}
Suppose $P$ admits a density $p\in L^1(\mathbb{R}^d)$.
Then $p$ can be recovered from its characteristic function by the inverse Fourier transform:
\[
p(x)
~=~
\frac{1}{(2\pi)^d}
\int_{\mathbb{R}^d}
e^{-i\,\omega^\top x}\,\varphi_P(\omega)\,d\omega,
\qquad \text{for a.e. }x\in\mathbb{R}^d.
\]
More generally, for any $f\in L^1(\mathbb{R}^d)$ with Fourier transform $\widehat f(\omega):=\int e^{i\,\omega^\top x}f(x)\,dx$,
\[
\int_{\mathbb{R}^d} f(x)\,P(dx)
~=~
\frac{1}{(2\pi)^d}
\int_{\mathbb{R}^d}\widehat f(\omega)\,\varphi_P(\omega)\,d\omega.
\]
Thus, the knowledge of $\varphi_P$ on $\mathbb{R}^d$ uniquely determines all integrals $\int f\,dP$, and hence the measure $P$ itself.
\end{proposition}

\begin{proof}[Proof (Fourier transform duality)]
The Fourier transform $\mathcal{F}:f\mapsto\widehat f$ is a linear bijection between $L^1$ and bounded continuous functions, satisfying the Plancherel identity $\int |f|^2 = (2\pi)^{-d}\int |\widehat f|^2$ for $f\in L^2$.
Since $e^{i\,\omega^\top x}$ is the kernel of this transform,
\begin{equation}
\begin{split}
\int f(x)\,P(dx)
~&=~
\int f(x)\!\left(\frac{1}{(2\pi)^d}\int e^{-i\,\omega^\top x}\,\varphi_P(\omega)\,d\omega\right)\!dx \\
~&=~
\frac{1}{(2\pi)^d}\int\widehat f(\omega)\,\varphi_P(\omega)\,d\omega.
\end{split}
\end{equation}
If $\varphi_P$ is known everywhere, the right-hand side gives $\int f\,dP$ for all $f\in L^1\cap L^2$, implying uniqueness of $P$.
\end{proof}

\vspace{0.3em}
\begin{proposition}[Lévy’s continuity theorem: weak convergence and uniqueness]
\label{prop:levy}
Let $\{P_n\}$ be a sequence of probability measures on $\mathbb{R}^d$ with characteristic functions $\varphi_{P_n}$.
Suppose that $\varphi_{P_n}(\omega)\to\varphi(\omega)$ for each $\omega$, and that the limit $\varphi$ is itself a characteristic function (i.e.\ positive definite and $\varphi(0)=1$).
Then $P_n\Rightarrow P$, where $P$ is the distribution with CF $\varphi$.
In particular, if $\varphi_P(\omega)\equiv\varphi_Q(\omega)$ for all $\omega$, then $P=Q$.
\end{proposition}

\begin{proof}[Proof (weak convergence via Fourier test functions)]
For any $f\in C_c^\infty(\mathbb{R}^d)$, its Fourier transform $\widehat f$ is rapidly decaying.
By Fubini and dominated convergence,
\begin{equation}
\begin{split}
& \int f(x)\,P_n(dx)
~=~
\frac{1}{(2\pi)^d}\int \widehat f(\omega)\,\varphi_{P_n}(\omega)\,d\omega
~\xrightarrow[n\to\infty]{}~ \\
& \frac{1}{(2\pi)^d}\int \widehat f(\omega)\,\varphi(\omega)\,d\omega
~=~
\int f(x)\,P(dx).
\end{split}
\end{equation}

Hence $\int f\,dP_n\to\int f\,dP$ for all $f\in C_c^\infty$, which implies $P_n\Rightarrow P$ by the Portmanteau theorem.
Taking $P_n=P$ and $\varphi_P=\varphi_Q$ gives the uniqueness $P=Q$.
\end{proof}

\vspace{0.5em}
\begin{proposition}[Bochner’s theorem: positive-definite functions as Fourier transforms of measures]
\label{prop:bochner}
A continuous function $\psi:\mathbb{R}^d\to\mathbb{C}$ with $\psi(0)=1$ is the characteristic function of some probability measure on $\mathbb{R}^d$ if and only if it is \emph{positive definite}, i.e.
for all $n\in\mathbb{N}$, all $\omega_1,\dots,\omega_n\in\mathbb{R}^d$, and all $c_1,\dots,c_n\in\mathbb{C}$,
\[
\sum_{i,j=1}^{n} c_i\,\overline{c_j}\,\psi(\omega_i-\omega_j)\ge 0.
\]
Moreover, for every finite nonnegative measure $\mu$ on $\mathbb{R}^d$, its Fourier transform
$\psi(\omega):=\int e^{\,i\,\omega^\top x}\,d\mu(x)$
is continuous and positive definite.
Thus, continuous positive-definite functions with $\psi(0)=1$ are in one-to-one correspondence with characteristic functions of probability measures.
\end{proposition}

\begin{proof}[Proof (spectral representation)]
For any finite positive measure $\mu$, $\psi(\omega)=\int e^{\,i\,\omega^\top x}\,d\mu(x)$ is continuous and satisfies the positive-definite inequality above by direct calculation.
Conversely, if $\psi$ is continuous and positive definite, by the classical Bochner--Khintchine theorem there exists a unique finite nonnegative measure $\mu$ such that $\psi$ is its Fourier transform.
Setting $\mu(\mathbb{R}^d)=1$ gives a probability measure $P=\mu$, with $\psi=\varphi_P$.
\end{proof}

\vspace{0.3em}
\begin{proposition}[Matching characteristic functions on $\mathbb{R}^d$ equals to matching the joint distribution]
\label{prop:cf-equivalence}
Let $P,Q$ be Borel probability measures on $\mathbb{R}^d$ with characteristic functions $\varphi_P,\varphi_Q$.
Define a nonnegative weighting function $w\in L^1(\mathbb{R}^d)$ satisfying $\mathrm{supp}(w)=\mathbb{R}^d$ and the frequency-domain distance
\[
\mathcal{D}_w^2(P,Q)
~:=~
\int_{\mathbb{R}^d} w(\omega)\,\big|\varphi_P(\omega)-\varphi_Q(\omega)\big|^2\,d\omega.
\]
Then the following are equivalent:
\begin{equation}
\begin{split}
\mathcal{D}_w(P,Q)=0
&\quad\Longleftrightarrow\quad
\varphi_P(\omega)=\varphi_Q(\omega)\ \forall\omega \\
&\quad\Longleftrightarrow\quad
P=Q.
\end{split}
\end{equation}

\end{proposition}

\begin{proof}[Proof (Lévy--Bochner synthesis)]
If $\mathcal{D}_w(P,Q)=0$, then $\varphi_P=\varphi_Q$ almost everywhere on $\{w>0\}$.
Since both $\varphi_P,\varphi_Q$ are continuous, equality extends to all $\omega\in\mathbb{R}^d$.
By Proposition~\ref{prop:levy}, $\varphi_P\equiv\varphi_Q$ implies $P=Q$. Conversely, if $P=Q$, then clearly $\mathcal{D}_w(P,Q)=0$.
Thus, equality of CFs on the full frequency domain is equivalent to equality of the underlying distributions.
\end{proof}

Intuitively, $D_w(P,Q)$ measures the squared discrepancy between characteristic
functions across all frequencies, weighted by $w$. When $w$ has full support, this
distance captures the complete distributional difference. We now connect this
frequency-domain view to kernel methods via Bochner's theorem.

\vspace{0.5em}
\begin{proposition}[Kernel formulation via Bochner’s theorem]
\label{prop:mmd-spectral}
Let $k(x,y)=\kappa(x-y)$ be a bounded, continuous, translation-invariant kernel with spectral measure $\Lambda$ satisfying
\[
\kappa(t)
~=~
\int_{\mathbb{R}^d} e^{\,i\,\omega^\top t}\,d\Lambda(\omega).
\]
Then the maximum mean discrepancy (MMD) between $P,Q$ in the RKHS of $k$ admits the spectral representation
\[
\mathrm{MMD}_k^2(P,Q)
~=~
\int_{\mathbb{R}^d}\big|\varphi_P(\omega)-\varphi_Q(\omega)\big|^2\,d\Lambda(\omega).
\]
Moreover, if the kernel is \emph{characteristic}—equivalently, the support of $\Lambda$ is all of $\mathbb{R}^d$—then $\mathrm{MMD}_k(P,Q)=0$ if and only if $P=Q$.
\end{proposition}

\begin{proof}[Proof (spectral integration)]
Expanding expectations and applying Bochner’s representation,

\begin{equation}
\begin{split}
\mathbb{E}\,k(X,Y)
~&=~
\int_{\mathbb{R}^d}\mathbb{E}\,e^{\,i\,\omega^\top(X-Y)}\,d\Lambda(\omega) \\
~&=~
\int_{\mathbb{R}^d}\varphi_P(\omega)\,\overline{\varphi_Q(\omega)}\,d\Lambda(\omega).
\end{split}
\end{equation}

Substituting into the definition
$\mathrm{MMD}_k^2(P,Q)=\mathbb{E}\,k(X,X')+\mathbb{E}\,k(Y,Y')-2\mathbb{E}\,k(X,Y)$
yields the stated formula.
If $\mathrm{MMD}_k(P,Q)=0$ while $\Lambda$ has full support, the integrand’s continuity forces $\varphi_P\equiv\varphi_Q$, which by Proposition~\ref{prop:levy} implies $P=Q$.
\end{proof}

\vspace{0.2em}
\paragraph{Summary and Connection to CFD.}
Propositions~2–6 collectively establish that a probability distribution is uniquely and completely determined by its characteristic function, and that matching characteristic functions over the entire frequency domain---either directly via a weighted $L^2$ distance $D_w$ or indirectly via a characteristic kernel MMD---is equivalent to matching the full joint distribution. The Fourier inversion theorem connects CFs to densities, Bochner's theorem connects positive-definite kernels to spectral measures, and Lévy's continuity theorem ensures that equality of CFs implies equality of distributions. Consequently, frequency-domain metrics such as our CFD provide a principled way to capture all moments and dependencies, beyond what marginal or covariance-based
criteria can express.

\section{Moments and Cumulants via Taylor Expansion of Characteristic Functions}
\label{sec:appendixb}

\subsection{Preliminaries and Notations}

Let $X=(X_1,\dots,X_d)\in\mathbb{R}^d$ be a random vector with law $P$.
Its \emph{characteristic function} (CF) is
\[
\varphi(\omega)\;=\;\mathbb{E}\!\left[e^{\,i\,\omega^\top X}\right],\qquad \omega=(\omega_1,\dots,\omega_d)\in\mathbb{R}^d,
\]
and the \emph{log-characteristic function} (log-CF) is $\psi(\omega)=\log \varphi(\omega)$ whenever $\varphi(\omega)\neq 0$.

\paragraph{Multi-index notation.}
A \emph{multi-index} is $\alpha=(\alpha_1,\dots,\alpha_d)\in\mathbb{N}_0^d$ with
$|\alpha|:=\sum_{j=1}^d \alpha_j$, $\alpha!:=\prod_{j=1}^d \alpha_j!$,
$\omega^\alpha:=\prod_{j=1}^d \omega_j^{\alpha_j}$, and $X^\alpha:=\prod_{j=1}^d X_j^{\alpha_j}$.
The mixed partial derivative is
\[
\partial^\alpha \;\coloneqq\; \frac{\partial^{|\alpha|}}{\partial \omega_1^{\alpha_1}\cdots \partial \omega_d^{\alpha_d}}.
\]

\vspace{-6mm}

\paragraph{Standing condition (finite moment of order $|\alpha|$).}
Throughout, whenever we speak of $\partial^\alpha\varphi(0)$ we assume $\mathbb{E}|X^\alpha|<\infty$.
This ensures, via dominated convergence, that differentiation and expectation can be interchanged.

\subsection{Limitations of Marginal-Only Matching}
Classical matching of \emph{univariate} moments (per-coordinate means/variances/skewness/kurtosis) aligns only the marginals $\{P_{X_j}\}_{j=1}^d$.
However, cross-variable dependence (pairwise, triple-wise, and beyond) lives in \emph{mixed} moments such as $\mathbb{E}[X_iX_j]$, $\mathbb{E}[X_iX_jX_k]$, etc.
Hence, marginal-only criteria cannot in general control the joint distribution.
We next show that the CF and its derivatives at the origin provide an exact, layered access to all mixed moments/cumulants, thereby capturing dependence.

\subsection{Taylor Coefficients and Mixed Moments/ Cumulants}

\begin{proposition}[Derivative--Moment Correspondence]
\label{prop:deriv-moment}
Suppose $\mathbb{E}|X^\alpha|<\infty$. Then
\[
\;\partial^\alpha \varphi(0)\;=\; i^{\,|\alpha|}\,\mathbb{E}\!\left[X^\alpha\right]\;
\]
for every multi-index $\alpha\in\mathbb{N}_0^d$.
\end{proposition}

\begin{proof}[Proof]
Write $e^{\,i\,\omega^\top X}=\prod_{j=1}^d e^{\,i\,\omega_jX_j}$ and differentiate.
Each differentiation w.r.t.\ $\omega_j$ contributes a factor $iX_j$.
Thus $\partial^\alpha e^{\,i\,\omega^\top X}=(iX_1)^{\alpha_1}\cdots(iX_d)^{\alpha_d}e^{\,i\,\omega^\top X}$.
Taking expectation and setting $\omega=0$ yields the claim, with the interchange of differentiation and expectation justified by $\mathbb{E}|X^\alpha|<\infty$.
\end{proof}
\medskip
\noindent\textbf{Low-order examples.}
For $\alpha=e_j$ (the $j$-th unit vector), $\partial^\alpha\varphi(0)=i\,\mathbb{E}[X_j]$ (means).
For $\alpha=e_j+e_k$, $\partial^\alpha\varphi(0)=-\,\mathbb{E}[X_jX_k]$ (mixed second moments).
For $\alpha=e_i+e_j+e_k$, $\partial^\alpha\varphi(0)=-\,i\,\mathbb{E}[X_iX_jX_k]$ (triple mixed moments).

\begin{proposition}[Multivariate Taylor expansion of $\varphi$ at $0$]
\label{cor:taylor-cf}
Under $\mathbb{E}|X^\alpha|<\infty$ for all $|\alpha|\le m$,
\begin{equation}
\begin{split}
\varphi(\omega)\;&=\;\sum_{|\alpha|\le m}\frac{1}{\alpha!}\,\partial^\alpha\varphi(0)\,\omega^\alpha \;+\; o(\|\omega\|^m) \\
\;&=\;\sum_{|\alpha|\le m}\frac{i^{\,|\alpha|}}{\alpha!}\,\mathbb{E}[X^\alpha]\,\omega^\alpha \;+\; o(\|\omega\|^m).
\end{split}
\end{equation}
\end{proposition}

Define $\psi(\omega)=\log \varphi(\omega)$ (well-defined near $0$ since $\varphi(0)=1$ and $\varphi$ is continuous).

\begin{definition}[Log-Characteristic Function and Mixed Cumulants]
\label{def:cumulant}
For a multi-index $\alpha\neq 0$ with $\mathbb{E}|X^\alpha|<\infty$, the \emph{mixed cumulant} is
\[
\;\kappa_\alpha \;\coloneqq\; i^{-\,|\alpha|}\,\partial^\alpha \psi(0)\;
\]
(with $\kappa_{\mathbf{0}}:=0$ by convention).
\end{definition}

\begin{proposition}[Derivative--Cumulant correspondence]
\label{thm:deriv-cumulant}
Assume the moments needed are finite so that $\psi$ is $|\alpha|$-times differentiable at $0$.
Then the coefficients of the multivariate Taylor series of $\psi$ at $0$ equal the mixed cumulants:

\begin{equation}
\begin{split}
&\psi(\omega)
=\sum_{|\alpha|\ge 1}\frac{1}{\alpha!}\,\partial^\alpha\psi(0)\,\omega^\alpha
=\sum_{|\alpha|\ge 1}\frac{i^{\,|\alpha|}}{\alpha!}\,\kappa_\alpha\,\omega^\alpha \\
&\quad\text{(convergent near $0$).}
\end{split}
\end{equation}

\end{proposition}

\begin{proof}[Proof]
By the chain rule, $\nabla\psi=\nabla\varphi/\varphi$ and
$\nabla^2\psi=(\nabla^2\varphi)/\varphi - (\nabla\varphi\,\nabla\varphi^\top)/\varphi^2$, etc.
Evaluating at $0$ and using Prop.~\ref{prop:deriv-moment}:

\begin{equation}
\begin{split}
&\nabla\psi(0)=i\,\mathbb{E}[X],\qquad \\
&\nabla^2\psi(0)=-\,\big(\mathbb{E}[XX^\top]-\mathbb{E}[X]\mathbb{E}[X]^\top\big)=-\,\mathrm{Cov}(X).
\end{split}
\end{equation}

Higher-order derivatives of $\psi$ yield the classical cumulant tensors (via the multivariate Faà di Bruno formula).
Collecting terms gives the stated Taylor expansion with coefficients $\partial^\alpha\psi(0)=i^{\,|\alpha|}\kappa_\alpha$.
\end{proof}
\medskip
\noindent\textbf{Interpretation and examples.}
$\kappa_{e_j}=\mathbb{E}[X_j]$ (means);
$\kappa_{e_j+e_k}=\mathrm{Cov}(X_j,X_k)$ (covariance);
third-order $\kappa_{e_i+e_j+e_k}$ measure non-Gaussian triple interactions (synergy/redundancy).
In general, \emph{cross-block independence} forces all mixed cumulants spanning the blocks to vanish, making cumulants a clean diagnostic of dependence.

\subsubsection*{Implications}
\begin{itemize}
\item Moments: $\partial^\alpha\varphi(0)=i^{|\alpha|}\mathbb{E}[X^\alpha]$ reveals \emph{all} mixed moments at each order $|\alpha|$.
\item Cumulants: $\partial^\alpha\psi(0)=i^{|\alpha|}\kappa_\alpha$ isolates genuine interactions (they are additive for independent sums and vanish across independent groups).
\item Hence, matching $\varphi$ (or $\psi$) near $0$ across \emph{all} multi-indices matches \emph{all} mixed moments/cumulants, aligning dependence at every order.
\end{itemize}

\section{Additional Analysis on PD-CFD}
\label{sec:appendixc}

Using the CUB-200-2011.\footnote{C. Wah, S. Branson, P. Welinder, P. Perona, and S. Belongie, \textit{Caltech-UCSD Birds-200-2011 Dataset}, Technical Report CNS-TR-2011-001, California Institute of Technology, 2011.} bird dataset illustrated in Fig.~\ref{fig:bird}, we evaluate how PD-CFD improves the
phase-focusing behavior of NCFM.\footnote{Wang et al., \textit{Dataset Distillation with Neural Characteristic Function: A Minmax Perspective}, in \textit{Proc. IEEE/ CVF Conf. Comput. Vis. Pattern Recognit. (CVPR)}, 2025, pp. 25570–25580. }. The CUB-200-2011 dataset contains extremely
fine-grained categories where discriminative information resides mainly in
high-frequency components such as feather textures. Therefore, the loss must be
sensitive to detailed structural variations.

We first revisit the original NCFM loss:
\begin{equation}
\begin{split}
&\mathrm{Chf}(t;f) 
= {} \alpha \left( \left|\Phi_{f(x)}(t) - \Phi_{\hat{f}(x)}(t)\right|^{2} \right) \\
& + (1-\alpha)\, \left|\Phi_{f(x)}(t)\right| \left|\Phi_{\hat{f}(x)}(t)\right| 
     \left( 1 - \cos\!\left( a_{f(x)}(t) - a_{\hat{f}(x)}(t) \right) \right) 
\end{split}
\label{eq:ncfm-extended}
\end{equation}
where $\Phi$ denotes the characteristic function magnitude and $a(\cdot)$ the
phase.

\begin{figure}[h]
    \centering
    \includegraphics[width=1\linewidth]{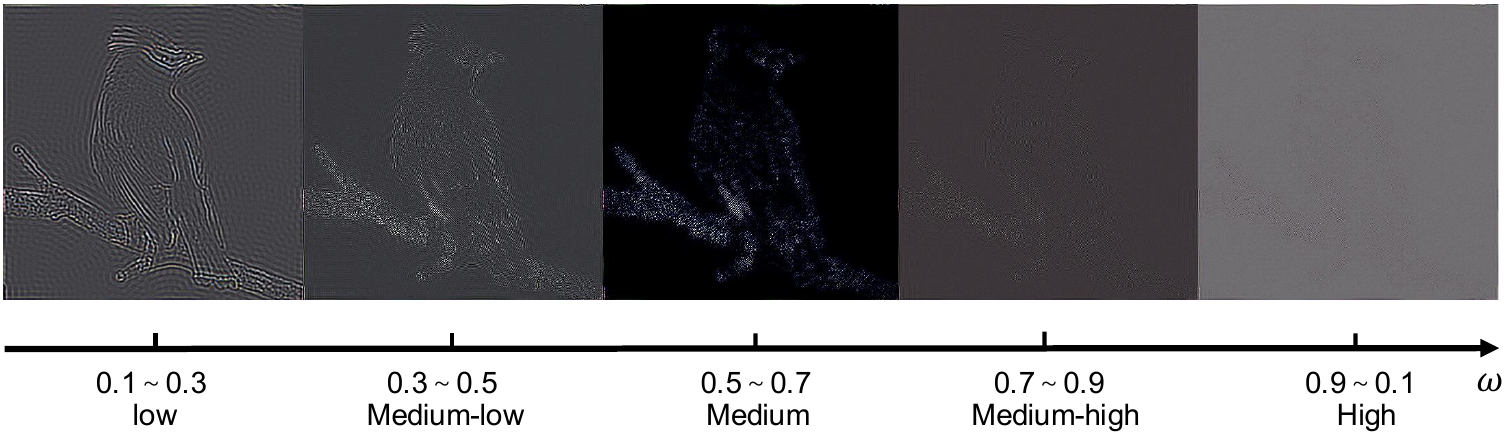}
    \caption{Phase information under different frequency band}
    \label{fig:phase}
\end{figure}

Although NCFM applies different hyperparameters to amplitude and phase, the
phase term is still tightly entangled with amplitude through the multiplicative
magnitude factor. As frequency increases, the amplitude decays, causing the
phase contribution to be increasingly suppressed. As a result, the mid- and
high-frequency phases---even those that still contain meaningful structural
information---are overwhelmed and incorrectly treated as noise. To illustrate
this behavior, we visualize phase differences across frequencies in Fig.\ref{fig:phase}: low-frequency
phase remains stable, mid-frequency phase begins to drift, and high-frequency
phase becomes dominated by boundary noise. However, some med- and med-high-frequency phase
components are still reliable and should not be discarded prematurely.

To restore these informative phase components, we extend the NCFM loss by adding
an explicit phase constraint defined for each sampled frequency $\omega \sim p(t)$ that is decoupled from amplitude :
\begin{equation}
\begin{split}
&\mathrm{Chf}(t;f) 
= {} \alpha \left( \left|\Phi_{f(x)}(t) - \Phi_{\hat{f}(x)}(t)\right|^{2} \right) \\
& + (1-\alpha)\, \left|\Phi_{f(x)}(t)\right| \left|\Phi_{\hat{f}(x)}(t)\right| 
     \left( 1 - \cos\!\left( a_{f(x)}(t) - a_{\hat{f}(x)}(t) \right) \right) \\
& + \frac{\lambda_{p}}{1 + \beta \|\omega\|^{2}}
  \left( \theta_{f(x)}(t) - \theta_{\hat{f}(x)}(t) \right)^{2},
\end{split}
\end{equation}

This additional phase term explicitly extracts the remaining reliable phase
information that is otherwise buried by amplitude attenuation in the original
NCFM formulation. In particular, it allows mid-frequency phase---which is still
semantically meaningful but degraded by the amplitude coupling---to be
effectively preserved.

\vspace{-4mm}

\begin{figure}[h]
    \centering
    \includegraphics[width=0.5\linewidth]{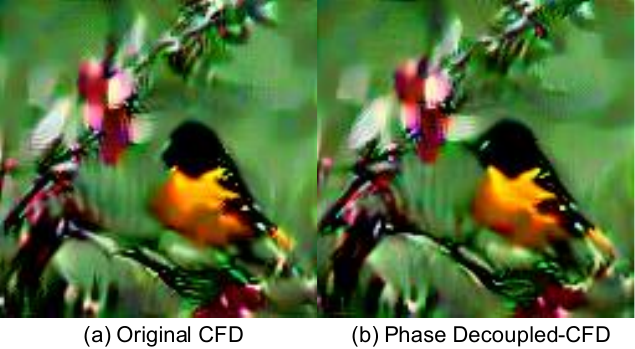}
    \caption{Comparison between generated data with original loss function and PD-CFD loss.}
    \label{fig:phase-conpare}
\end{figure}

\vspace{-4mm}

We verify the effect of this decoupled phase constraint under downstream classification accuracy and results are illustrated in Table.~\ref{tab:phase_ablation} (see Fig.~\ref{fig:phase-conpare} for an example of the synthetic image). The results match 
our expectation: when $\lambda_{p}=0.3$, the added phase regularization 
significantly strengthens high-frequency alignment, especially for the feather 
and edge regions in the CUB images.

\begin{table}[htbp]
\centering
\caption{Effect of phase regularization $\lambda_{p}$ on accuracy.}
\vspace{-2mm}
\begin{tabular}{ccc}
\toprule
\textbf{$\lambda_{p}$} & \textbf{Accuracy (\%)} & \textbf{Improvement} \\
\midrule
0.0 & 23.80 & 0 \\
0.1 & 24.15 & 1.47\% \\
0.2 & 25.20 & 5.88\% \\
0.3 & 28.35 & 19.12\% \\
\bottomrule
\end{tabular}
\label{tab:phase_ablation}
\end{table}

A similar phenomenon is observed on the DTD and RESISC-45 datasets(illustrated in Fig.~\ref{fig:dtd} Fig.~\ref{fig:resisc45} respectively).
The DTD texture dataset contains a large amount of rapidly oscillating material patterns characterized by abrupt edge transitions, fine-scale boundaries, and dense local contour variations. Likewise, the RESISC-45 remote-sensing dataset exhibits substantial geometric complexity:
land-cover boundaries, building edges, road networks, rooftop outlines, and other high-frequency landscape structures that vary sharply across classes. These discriminative details are dominated by high-order moment information in the frequency domain. Traditional metrics such as MSE or CE are inherently insensitive to such structural discontinuities---they primarily respond to low-order statistics and smooth variations, and therefore fail to capture the fine-grained distributional differences arising from high-frequency
contours and edge transitions.

In contrast, our PD-CFD formulation introduced earlier
can naturally recover these informative distributional discrepancies.
By removing the amplitude-induced suppression of mid- and high-frequency
phase, PD-CFD preserves the remaining reliable phase components that
encode exactly these texture- and edge-based variations. Consequently,
FAST maintains strong performance even on challenging datasets such as
DTD and RESISC-45, where structural information is dominated by, high-frequency features and where competing baselines degrade
substantially.

\vspace{-8mm}

\begin{figure}[b]
    \centering
    \includegraphics[width=0.8\linewidth]{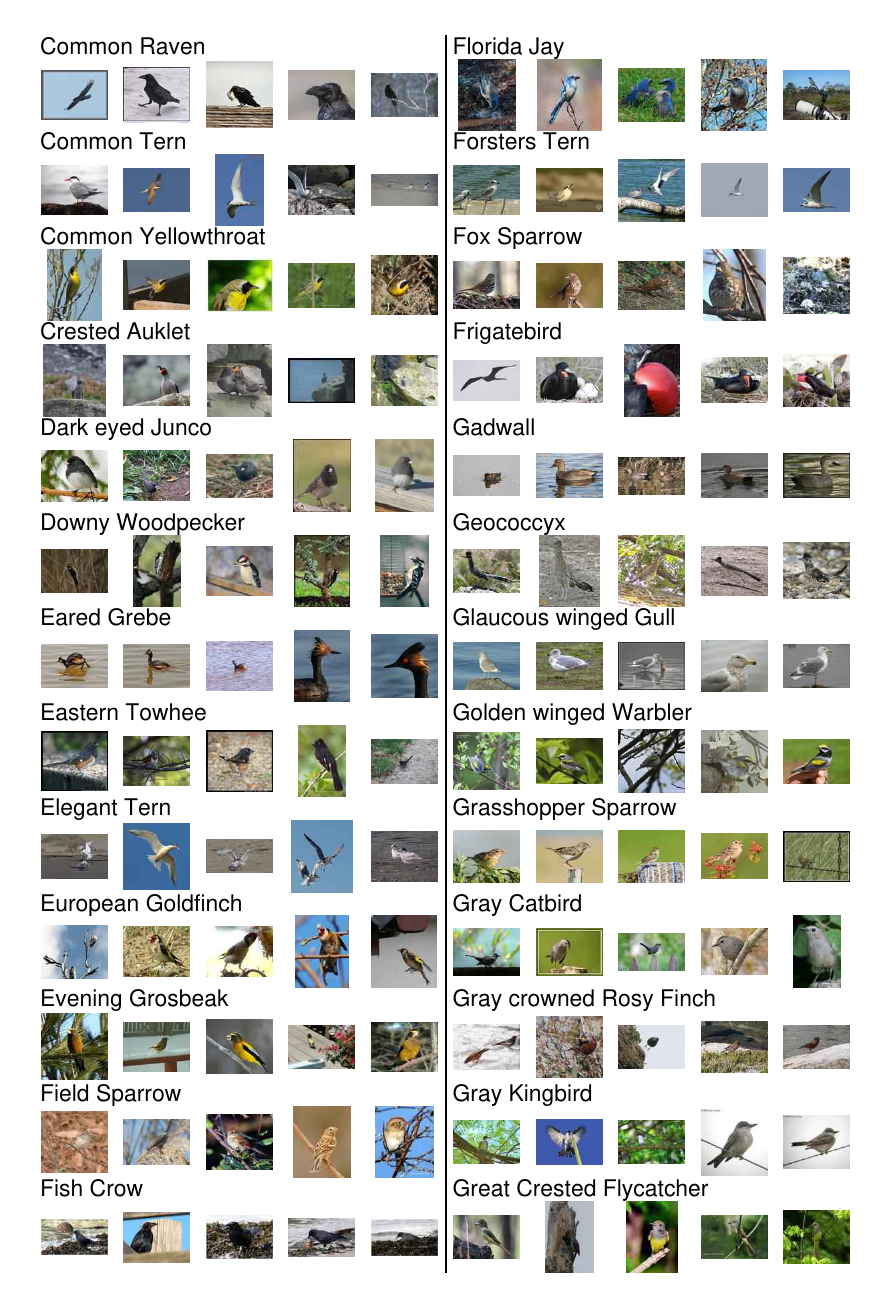}
    \caption{CUB-200-2011 Bird Dataset}
    \label{fig:bird}
\end{figure}

\begin{figure}[b]
    \centering
    \includegraphics[width=0.8\linewidth]{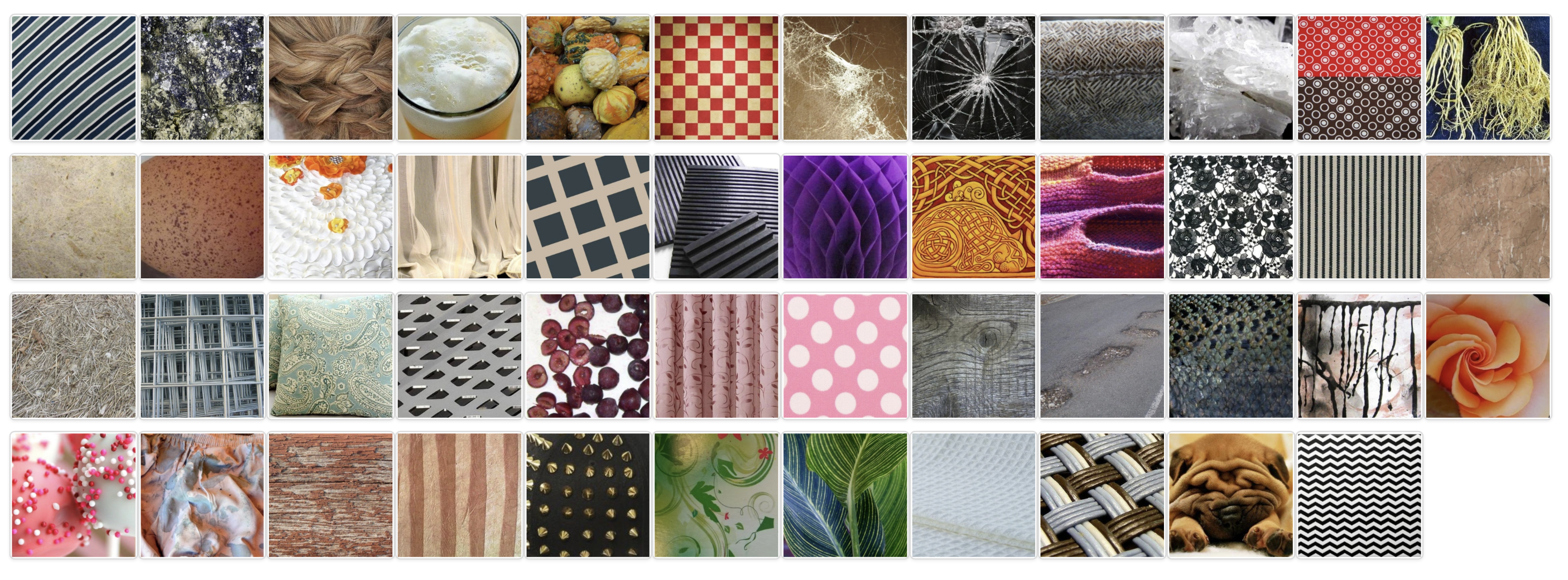}
    \caption{DTD Texture Dataset}
    \label{fig:dtd}
\end{figure}

\begin{figure}[b]
    \centering
    \includegraphics[width=0.8\linewidth]{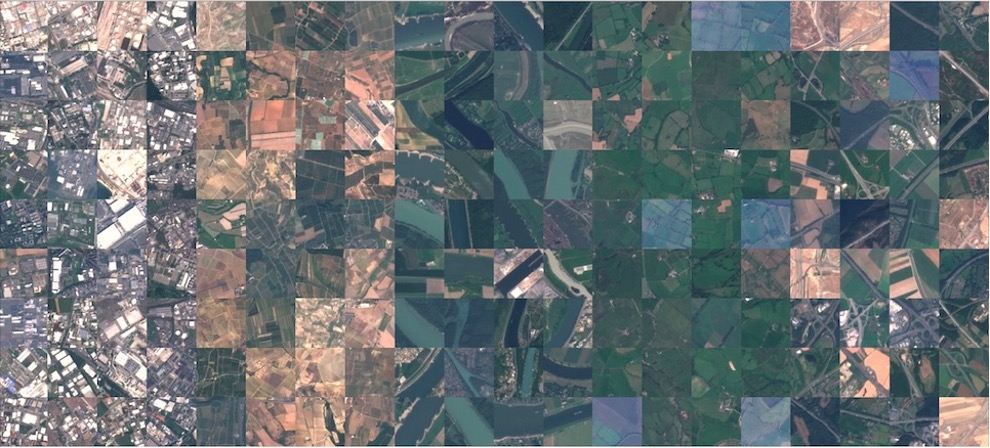}
    \caption{RESISC-45 Remote Sensing Dataset}
    \label{fig:resisc45}
\end{figure}

\section{Experiments on Language Tasks}
\label{sec:appendixd}

\subsection{Setup}
\noindent To further evaluate the generalization capability of FAST on LLM datasets, we conduct experiments on the Alpaca instruction-following dataset. Following our main setup, we adopt LLaMA2-7B as the backbone and apply LoRA-based fine-tuning using core-sets sampled at retention rates of 10\%, 20\%, and 30\%. For evaluation, we report performance on the MMLU benchmark, which spans 57 diverse subjects and serves as a rigorous protocol for assessing broad knowledge and reasoning abilities. We compare FAST against the NMS baseline\footnote{Boran Zhao et al., \textit{NMS:
Efficient Edge DNN Training via Near-Memory Sampling on Manifolds}, arXiv preprint arXiv:2508.02313, 2025.}, the current SOTA method, under identical experimental settings.

\subsection{Results}
\begin{figure*}[t]
\centering
\includegraphics[width=\textwidth]{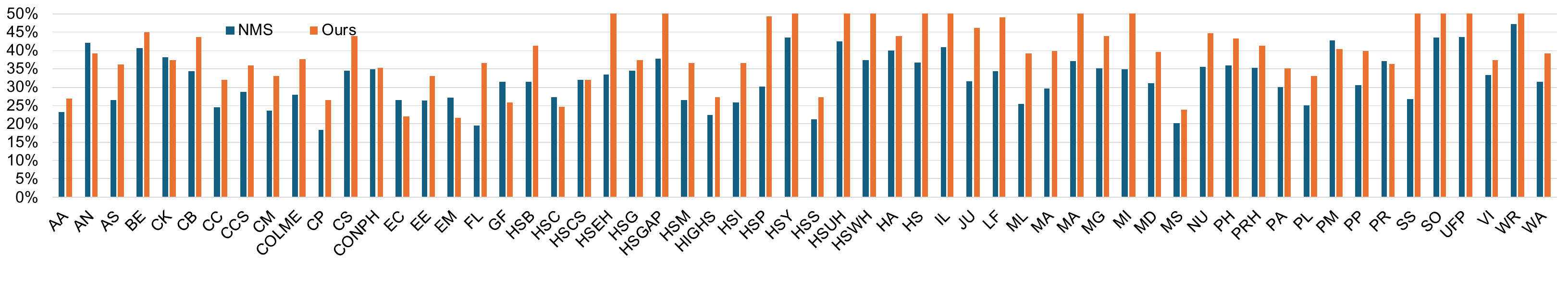}\\[-4pt]
\footnotesize (a) Keep ratio 10\% \\[6pt]

\includegraphics[width=\textwidth]{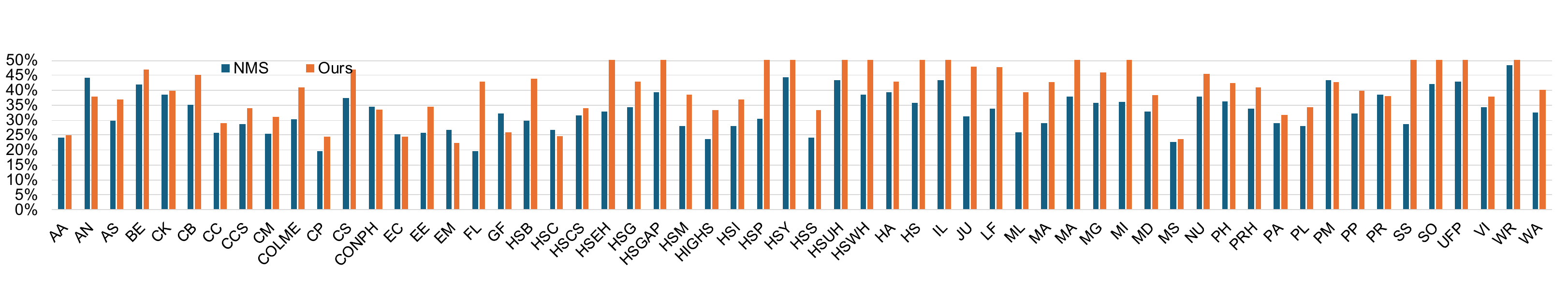}\\[-4pt]
\footnotesize (b) Keep ratio 20\% \\[6pt]

\includegraphics[width=\textwidth]{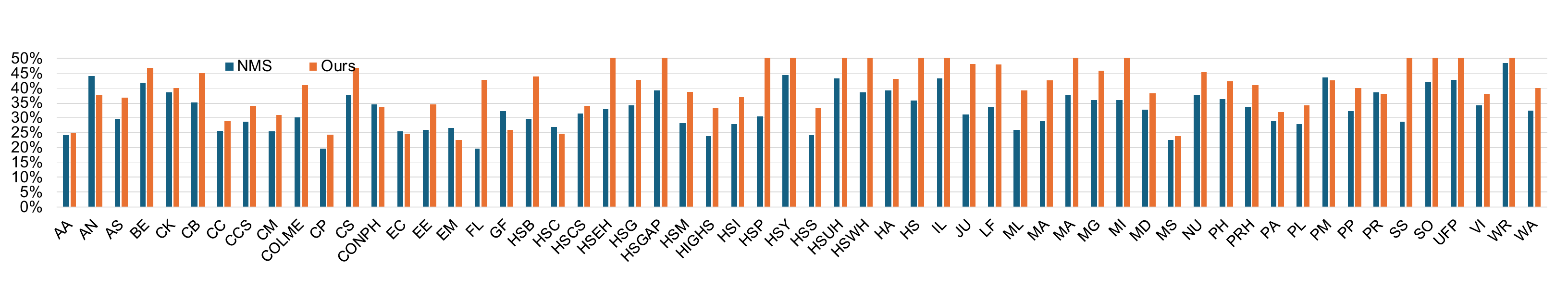}\\[-4pt]
\footnotesize (c) Keep ratio 30\%

\caption{MMLU accuracy comparison at different keep ratios.}
\label{fig:mmlu}
\end{figure*}

\noindent The MMLU results are shown in Fig.~\ref{fig:mmlu}. Across all keep ratios, FAST consistently outperforms the NMS baseline. While NMS attains an average accuracy of approximately 33\%, FAST increases the performance to 39\%, corresponding to a relative improvement of about 18\%. These results indicate that FAST is highly effective at selecting and preserving semantically informative instances. Moreover, the strong performance gains suggest that high-level semantic structure can be retained through spectral-graph-based distribution alignment alone, without explicit reliance on neural feature extractors. By leveraging geometric and frequency-domain signals, FAST preserves the underlying semantic neighborhood relations required for downstream reasoning. This demonstrates the robustness and broad generalization ability of FAST, even on tasks that rely heavily on semantic consistency.

\end{document}